\documentclass{article}

\usepackage[numbers]{natbib}

\usepackage{hyperref}
\usepackage{tabularx}
\usepackage[dvipsnames,svgnames]{xcolor}
\usepackage{graphicx}
\usepackage{booktabs}       %
\usepackage{amsmath,amssymb,amsthm}
\usepackage{subcaption}
\usepackage{algorithm} 
\usepackage{algpseudocode} 
\usepackage{capt-of}        %
\usepackage{bm}             %
\usepackage{enumitem}       %
\usepackage{tikz,calc}
\usepackage{cleveref}       %
\crefformat{footnote}{#2\footnotemark[#1]#3}

\theoremstyle{remark}
\newtheorem{remark}{Remark}

\usetikzlibrary{shadows}
\usetikzlibrary{calc}
\usetikzlibrary{trees}
\usetikzlibrary{mindmap}
\usetikzlibrary{shapes.callouts}
\usetikzlibrary{positioning}
\usetikzlibrary{decorations.text}
\usetikzlibrary{decorations.pathmorphing}
\usetikzlibrary{shapes.arrows}
\usetikzlibrary{shapes.multipart}
\usetikzlibrary{backgrounds}
\usetikzlibrary{tikzmark}
\usetikzlibrary{arrows.meta}
\usetikzlibrary{matrix}
\tikzstyle{every picture}+=[remember picture]

\DeclareBoldMathCommand{\bbeta}{\beta}
\DeclareBoldMathCommand{\btheta}{\theta}
\DeclareBoldMathCommand{\latentdirection}{\delta}
\DeclareBoldMathCommand{\bmu}{\mu}
\DeclareBoldMathCommand{\a}{a}
\DeclareBoldMathCommand{\e}{e}
\DeclareBoldMathCommand{\x}{x}
\DeclareBoldMathCommand{\v}{v}
\DeclareBoldMathCommand{\z}{z}

\newcommand{\xcontrast}{\x_\textrm{c}}
\newcommand{\zcontrast}{\z_\textrm{c}}
\newcommand{\xboundary}{\x_\textrm{b}}
\newcommand{\xboundaryj}[1]{\x_{\textrm{b},#1}}
\newcommand{\zboundary}{\z_\textrm{b}}
\newcommand{\estbeta}{\hat{\bbeta}}
\newcommand{\esttheta}{\hat{\btheta}}
\newcommand{\intercept}{\hat{\beta}_0}

\newcommand{\dbasimple}{\textsc{DBA-Tab}}
\newcommand{\dbafull}{\textsc{DBA-Att}}
\newcommand{\limefull}{\textsc{LIME-Att}}
\newcommand{\globalsurrogate}{\textsc{Global}}
\newcommand{\reals}{\mathbb{R}}
\newcommand{\inputspace}{\mathcal{X}}
\newcommand{\labelspace}{\mathcal{Y}}
\newcommand{\latentspace}{\mathcal{Z}}
\newcommand{\attributespace}{\mathcal{A}}
\newcommand{\rgrid}{\mathcal{R}}
\newcommand{\PL}{\textrm{PL}}   %
\newcommand{\PW}{\textrm{PW}}   %
\newcommand{\SL}{\textrm{SL}}   %
\newcommand{\SW}{\textrm{SW}}   %
\newcommand{\C}{\textrm{C}}     %
\newcommand{\sPL}{\widetilde{\PL}}  %
\newcommand{\sPW}{\widetilde{\PW}}  %
\newcommand{\sSL}{\widetilde{\SL}}  %
\newcommand{\sSW}{\widetilde{\SW}}  %
\newcommand{\sC}{\widetilde{\C}}    %
\newcommand{\normaldist}{\mathcal{N}}    %
\newcommand{\empmean}{\hat{\bmu}}    %
\newcommand{\empSigma}{\hat{\Sigma}}  %

\DeclareMathOperator{\diag}{diag}

\title{Explaining Predictions by Approximating\\the Local Decision
Boundary}

\author{Georgios Vlassopoulos\\
Jheronimus Academy of Data Science\\
\texttt{georgiosvlassopoulos@gmail.com}%
\and
Tim van Erven\\
University of Amsterdam\\
\texttt{tim@timvanerven.nl}%
\and
Henry Brighton\\
Tilburg University\\
\texttt{brighton@uvt.nl}%
\and
Vlado Menkovski\\
Eindhoven University of Technology\\
\texttt{v.menkovski@tue.nl}%
}

\begin{document}

\maketitle

\begin{abstract}
  Constructing accurate model-agnostic explanations for opaque machine
  learning models remains a challenging task. Classification models for
  high-dimensional data, like images, are often inherently complex. To
  reduce this complexity, individual predictions may be explained
  locally, either in terms of a simpler local surrogate model or by
  communicating how the predictions contrast with those of another
  class. However, existing approaches still fall short in the following
  ways: a)~they measure locality using a (Euclidean) metric that is not
  meaningful for non-linear high-dimensional data; or b) they do not
  attempt to explain the decision boundary, which is the most relevant
  characteristic of classifiers that are optimized for classification
  accuracy; or c) they do not give the user any freedom in specifying
  attributes that are meaningful to them. We address these issues in a
  new procedure for local decision boundary approximation (DBA). To
  construct a meaningful metric, we train a variational autoencoder to
  learn a Euclidean latent space of encoded data representations. We
  impose interpretability by exploiting attribute annotations to map the
  latent space to attributes that are meaningful to the user. A
  difficulty in evaluating explainability approaches is the lack of a
  ground truth. We address this by introducing a new benchmark data set
  with artificially generated Iris images, and showing that we can
  recover the latent attributes that locally determine the class. We
  further evaluate our approach on tabular data and on the CelebA image
  data set.
\end{abstract}

\section{Introduction}

Over the last few years, explaining opaque machine learning (ML) models
has become a topic of increasing attention
\cite{BarredoEtAl2019-taxonomy1,taxonomy3,taxonomy4,taxonomy5,taxonomy6}. This
attention arises from multiple needs of ML users, such as ensuring model
\emph{trustworthiness}, detecting and removing unwanted biases
(\emph{fairness}) and understanding \emph{causal relationships}
\cite{BarredoEtAl2019-taxonomy1}. For many of these needs, it is crucial that
explanations are able to identify and communicate which properties of
the input are the most important for the model's predictions. These
properties can be identified either by modifying the underlying
mechanism of the model in a trade-off with predictive accuracy, or by
designing algorithms that explain the model's behavior post-hoc after
training. For the latter, one may take into account the model's internal
mechanisms and develop frameworks for a specific category of models
(e.g.\ for neural
networks \cite{LRP,TCAV,CAM} or random forests \cite{forestfloor}), or
take a \emph{model-agnostic} approach that can explain many different
types of models \cite{LIME,Shapp,grex,distill_n_compare,sensitivity,CEM}.

Predictive models are often complex and controlled by 
many parameters that cannot all be communicated to a user. This
complexity can be reduced by explaining only the model's
behavior in a \emph{local} region of the feature space near a given
input $\x_0$. Local model-agnostic explanations may be
\emph{example-based}, by simulating a data point $\xcontrast$ from a
different class that is close to $\x_0$ \cite{gspheres,CEM}. This has the
drawback that it is left to the user to identify the relevant
differences between $\xcontrast$ and $\x_0$, which may be difficult for
high-dimensional data like images, especially when $\xcontrast$ and
$\x_0$ are close. An alternative is to approximate the model locally by
an interpretable \emph{surrogate model}, which is usually a linear
model, a set of decision rules or a decision tree
\cite{LIME,anchor}.

We propose three criteria that local model-agnostic
explanations should satisfy. First, locality needs to be measured by a
\emph{metric} that is \emph{meaningful} for the data at hand. For
unstructured low-dimensional data without strong correlations between
the features, it may be very reasonable to use Euclidean distance, but
for highly structured data we typically need a more refined measure. For
instance, the difference in pixel values between two images will be
affected greatly by the slight movement of an object, by a variation in
lighting or by adding minor noise, without users perceiving any relevant
difference in content. Lack of an appropriate metric prevents some
explainability methods from scaling up to image data
\cite{LocalSurrogate} and is not even fully avoided by some methods that
do apply to image data \cite{CEM-MAF}.

Second, we posit that explanations of classifiers should depend solely
on the \emph{decision boundary} between classes, and not on any class
probability estimates the classifiers may output in addition. Since
classifiers are typically optimized for prediction accuracy and not for
probability estimation, they often output probabilities that are
uncalibrated
\cite{GuoEtAl2017-NNCalibration,Niculescu-MizilEtAl2005-OldSchoolCalibration}
and therefore lack interpretability. Moreover, users are directly
affected by the decision boundary and not by probabilities: someone who
is refused a mortgage may not care whether a better credit score could
have reduced their estimated probability of defaulting on payment from
0.62 to 0.61, but they likely would care what would have changed the
bank's decision. Finally, output probabilities may average the influence
of multiple nearby parts of the decision boundary together, and thereby
fail to represent any of these parts accurately, as shown by
\citet{LocalSurrogate}.

Third, a consensus appears to be forming in the literature that
explanations need to be tailored to the target audience and context
\cite{BarredoEtAl2019-taxonomy1,taxonomy6}: an ML developer
who is trying to fix classification errors should not be given the same
output as an authority assessing compliance with fairness regulations.
Consequently, there should be a way for the user to \emph{specify
attributes} of the data that are interpretable and relevant to them.
When the user is an ML developer, such attributes may sometimes take a
convenient form that can be automatically extracted, like for instance
superpixels in an image. But in general it may be required to invest
additional resources into learning the concepts that a given type of
user can understand, for instance by collecting additional data
annotations.

\paragraph{New Method}

We introduce the first model-agnostic method to explain binary
classifiers that meets all three of these criteria. The procedure
performs local decision boundary approximation (DBA) by searching for
the closest decision boundary point $\xboundary$ to $\x_0$, and
generating a sample of artificial data points around it which are
labeled by the model that we want to explain. A linear surrogate model
is then fit to this sample. We interpret the coefficients of this linear
model as the direction of minimal change to switch the class of $\x_0$.
Instead of measuring distance in the input space, we train a variational
autoencoder (VAE) \cite{KingmaWelling2013} to learn a Euclidean latent
space of encoded data representations, and measure distance in this
latent space. We also use attribute annotations to learn mappings from
the latent space to user-specified attributes, and fit the linear
surrogate in terms of the attributes. We restrict the search for
$\xboundary$ to the data manifold by performing bisecting line searches
between $\x_0$ and a selection of nearby training data points from the
opposite class.
This has the additional benefit that the run-time of the search
procedure (almost) does not depend on the dimension. Both the line
searches and the sampling step take place in the latent space learned by
the VAE. Our main contributions lie in the new search and sampling
procedures and the way they are combined with a VAE and user-specified
attributes. We further adjust the training procedure of the VAE to favor
preservation of class probabilities, which can have a significant
effect.

\paragraph{New Evaluation Data}

A difficulty in evaluating explainability approaches is the lack of a
ground truth. To remedy this, we introduce the Artificial Iris (AIris)
data set, with simulated images of flowers that can be used as a
benchmark for explainability methods. We label the classes based on two
hyperplanes defined by latent parameters of the flowers like petal
length and sepal width. We then show that our method is able to recover
the true coefficients of the hyperplanes with high accuracy from the
predictions of a convolutional neural network (CNN) that is trained to
classify near-perfectly.

\paragraph{Related Work}

Existing methods for local model-agnostic explanation include LIME
\cite{LIME}, kernel SHAP \cite{Shapp}, LORE \cite{LORE} and MAPLE
\cite{MAPLE}. Among these, kernel SHAP, LORE and MAPLE are only for
tabular data, and therefore do not directly address our first and third
criteria of using a meaningful metric and allowing the user to specify
interpretable attributes. In addition, none of these methods aims to
explain the decision boundary (our second criterion): LIME and SHAP
sample around $\x_0$ instead of near the decision boundary. The same holds
for the way LORE generates samples from the same class as $\x_0$. MAPLE
weights training points by their probability of being in the same class as
$\x_0$, which again leads to high weights near $\x_0$ but not on the
opposite side of the decision boundary. For LIME, the sampling procedure
and the surrogate model can be specified in terms of abstract binary
attributes of $\x_0$, for which the authors propose specific choices
(e.g.\ superpixels for images), but which can in principle be specified
by the user (thus satisfying our third criterion).

Our use of a latent space for model-agnostic explanation is not new. For
instance, word embeddings have been widely used for natural language
data to extend the applicability of tabular explanation methods
\cite{LIME,L2X}, ALIME \cite{ALIME} uses a denoising autoencoder,
CEM-MAF \cite{CEM-MAF} uses the latent space of a VAE or generative
adversarial network (GAN), and ABELE \cite{ABELE} uses an adversarial
autoencoder. The difference with these prior works is that none of them
tries to explain the decision boundary directly: ALIME is an extension
of LIME, CEM-MAF generates a contrastive example $\xcontrast$ from the
opposite class that is near to $\x_0$ but need not be close to the
decision boundary, and ABELE uses the sampling procedure from LORE.

Existing methods that do attempt to explain the decision boundary
directly are L2X \cite{L2X} and LocalSurrogate
\cite{LocalSurrogate,gspheres}. But again these are both for tabular
data only. L2X does incorporate a meaningful `metric' based on mutual
information between $\x_0$ and its explanation, but it can only output
explanations that consist of a sparse subset of elements of $\x_0$, so
it fails our third criterion of incorporating user-specified attributes.
LocalSurrogate is similar in spirit to our DBA procedure, but attempts
to sample densely in a sphere, which is not feasible beyond very small
dimensions.

Finally, we mention ANCHORS \cite{anchor}, which produces local decision
rules that are consistent with the decision boundary, and
TCAV \cite{TCAV}, which uses attribute annotations to learn a mapping from the
internal state of a neural network to user-defined attributes.

\paragraph{Outline}

\enlargethispage{\baselineskip}

In the next section we describe our new DBA method.
Section~\ref{sec:AIris} then introduces the Artificial Iris data set and
presents experiments comparing DBA to a variant of LIME and to CEM-MAF.
Section~\ref{sec:celeba} extends the evaluation to the CelebA data set
\cite{celeba} with real-world celebrity images. We show there that
blurriness of images and open mouths improperly affect classifications.
Finally, we conclude with a summary and discussion of the requirement to
apply our approach. Details and additional experiments, including a
comparison between DBA, LIME and MAPLE on tabular data from several
standard UCI data sets, are postponed to the Appendix.

\section{Method: Local Decision Boundary Approximation}
\label{sec:dba}

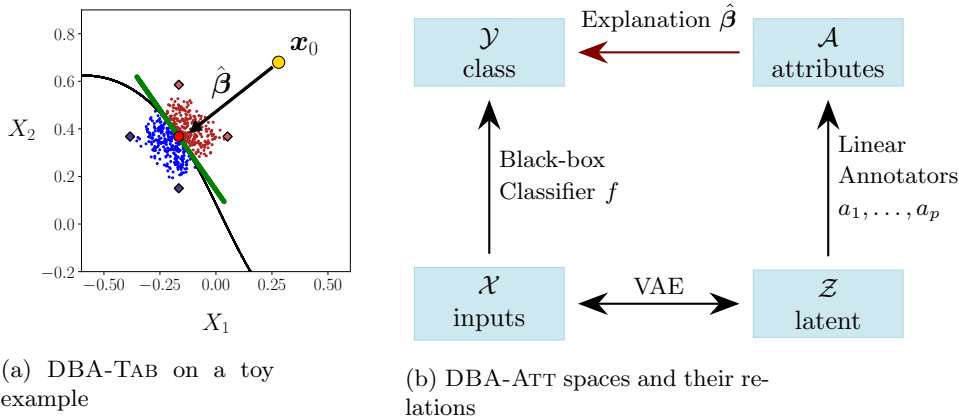
\begin{figure}[tbp]
    \begin{subfigure}{0.3\textwidth}
        \includegraphics[height=4.5cm]{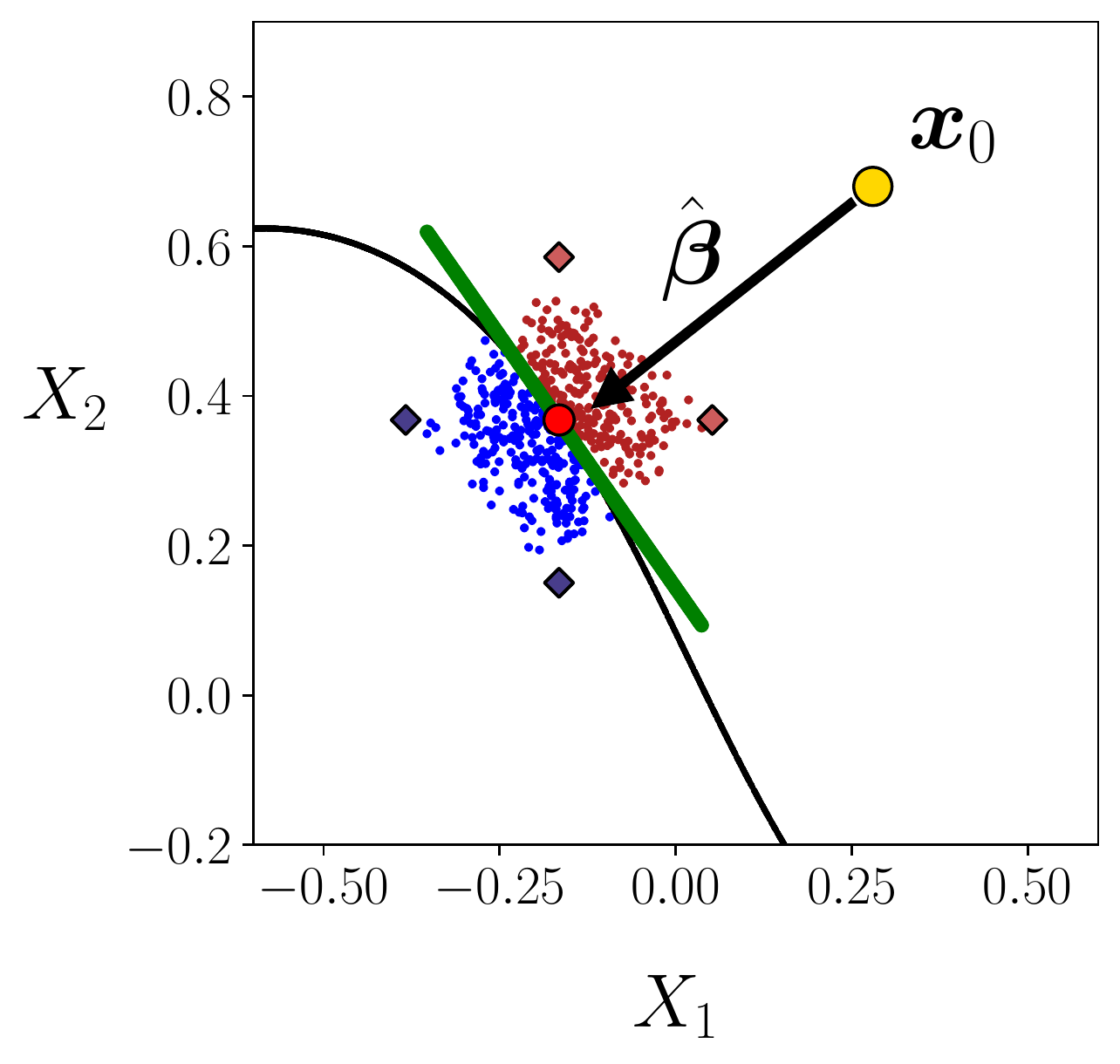}
        \caption{\dbasimple{} on a toy example}
        \label{fig:dba_toy}
    \end{subfigure} %
    \hspace{1.5cm}
    \begin{subfigure}{0.4\textwidth}
        \begin{tikzpicture}[
          space/.style={align=center,rectangle,fill=LightBlue!60,draw=LightBlue,text
          width=5em},
          arrowstyle/.style={thick,>={Stealth[length=4mm]}},
          shorten >=0.5em,
          shorten <=0.5em,
          node distance=2cm]
        \matrix[ampersand replacement=\&,row sep=24mm,column sep=25mm] {
          \node [space] (class) {$\labelspace$ \\ class};
          \& \node [space] (attributes) {$\attributespace$ \\ attributes};
          \\
          \node [space] (inputs) {$\inputspace$ \\ inputs};
          \& \node [space] (latent) {$\latentspace$ \\ latent};
          \\
        };
       \path[->,arrowstyle,color=Maroon] (attributes) edge node
       [above=1ex,color=black]
       {\small{Explanation $\estbeta$}} (class);
       \path[->,arrowstyle] (inputs) edge node [right,text width=6em]
       {\small{Black-box Classifier $f$}} (class);
       \path[<->,arrowstyle] (inputs) edge node [above]
       {\small{VAE}} (latent);
       \path[->,arrowstyle] (latent) edge node [right,text width=4em]
       {\small{Linear\\ Annotators $a_1,\ldots,a_p$}} (attributes);
        \end{tikzpicture}
        \caption{\dbafull{} spaces and their relations}
        \label{fig:attdba_scheme}
    \end{subfigure}
    \caption{Illustration of the \dbasimple{} and \dbafull{} procedures}
    \label{fig:illustration}
\end{figure}

Let $f : \inputspace \to \labelspace = \{-1,+1\}$ be the binary classifier, whose
decision for input $\x_0 \in \inputspace \subset \reals^d$ is to be
explained. Suppose also that the training data $D =
\big((\x_1,y_1),\ldots,(\x_n,y_n)\big)$ on which $f$ has been trained are
still available, with $\x_i \in \inputspace$ and $y_i \in \labelspace$. We
first present a simplified version of our method: \dbasimple{}, which is
suitable for tabular data. This simplified version applies if the inputs
$\x$ consist of $d$ features that are already interpretable to the user,
so we do not need a VAE or separate interpretable attributes. It further
requires that the features have been suitably standardized and
decorrelated to make Euclidean distance an appropriate metric for
$\inputspace$. \dbasimple{} consists of the following steps, which are
illustrated in Figure~\ref{fig:dba_toy}:
\begin{enumerate}
\item \emph{Detection:} Search along the manifold of the training data
$D$ to find the point $\xboundary$ on the decision boundary that is
closest to $\x_0$.
\end{enumerate}
This is implemented by selecting the $k$ closest points to $\x_0$ from
$D$ that are from the opposite class. For each selected point
$(\x_j,y_j)$, we then perform a line search using the bisection method
to find a point $\xboundaryj{j}$ on the line segment between
$\x_0$ and $\x_j$ that is on the decision boundary of $f$. We call
$\x_j$ the \emph{corresponding bisected point} for $\xboundaryj{j}$.
Finally, among these decision boundary points
$\xboundaryj{1},\ldots,\xboundaryj{k}$, we choose $\xboundary$ to be the
point that is closest to~$\x_0$. A notable advantage of this approach is
computational efficiency: instead of searching in $d$ dimensions for a
boundary of unknown shape, it only needs to perform $k$ one-dimensional
line searches.
\begin{enumerate}[resume]
\item \emph{Simulation:} Randomly generate $m$ points near $\xboundary$
on both sides of the decision boundary of~$f$ and label them with $f$ to
obtain a sample $S = \big((\x_1,y_1),\ldots,(\x_m,y_m)\big)$.
\end{enumerate}
A natural first idea would be to sample densely from a sphere around
$\xboundary$, but then the curse of dimensionality
\cite{HastieTibshiraniFriedman2009} would imply that the sample size $m$
would need to grow exponentially with $d$, so this is hopeless except
for very small $d$. Instead, we generate a sample that contains
variation in all the feature directions: for each feature
$j=1,\ldots,d$, we first create two vertices $\v_{j,-1}$ and $\v_{j,+1}$
that are equal to $\xboundary$ except that they respectively decrease
and increase the value of feature $j$ by an amount~$\alpha$. We set
$\alpha = r \|\xboundary - \x_0\|$ proportional to the distance between
$\xboundary$ and $\x_0$ with a proportionality parameter $r > 0$ that is
specified below. Each point $(\x,y) \in S$ is then sampled independently
from the convex hull of the vertices $\v_{1,\pm 1},\ldots,\v_{d,\pm 1}$
by drawing random weights $w_{j,\pm 1}$ uniformly at random from the
$2d$-dimensional probability simplex, and setting $\x = \sum_{j=1}^d
w_{j,-1} \v_{j,-1} + \sum_{j=1}^d w_{j,+1} \v_{j,+1}$. We note that this
approach does not sample uniformly from the convex hull of the vertices,
but instead induces approximate sparsity in the samples, because the
influences of $\v_{j,-1}$ and $\v_{j,+1}$ approximately cancel each
other out when their weights $w_{j,-1}$ and $w_{j,+1}$ are similar. Our
use of (approximate) sparsity bears a resemblance to LIME \cite{LIME},
which samples sparse perturbations of $\x_0$.
\begin{enumerate}[resume]
  \item \emph{Explanation:} Fit a linear surrogate model $g(\x) =
  \x^\top \estbeta + \intercept$ on the sample $S$ using (unpenalized)
  logistic regression.
\end{enumerate}
The parameters of the method are $k, m$ and $r$. We tune $r$
automatically from a grid $\rgrid$ of possible values by choosing the
value for which the resulting direction $\estbeta$ minimizes the
distance from $\x_0$ to the decision boundary.\footnote{The distance to
the decision boundary is again determined by the bisection method; this
time on the line segment between $\x_0$ and $\x' = \x_0 - f(\x_0)\gamma
\estbeta /\|\estbeta\|$, where we take $\gamma$ large enough that $\x'$
is always on the other side of the decision boundary. Specifically, we
use $\gamma = \|\x_0 - \xboundary\| + 0.1$ in all our experiments.}
The output of the algorithm is the linear surrogate model $g$. Its most
important component is the coefficient vector $\estbeta$, which is the
normal vector to the decision boundary of $g$ and can be interpreted as
the direction of minimal change to switch the class of $\x_0$, as
illustrated in Figure~\ref{fig:dba_toy}. See Appendix~\ref{app:dba} for
experiments illustrating the behavior of \dbasimple{}.

\subsection{Extension to Structured High-dimensional Data}
\label{sec:dbafull}

\dbasimple{} is not suitable for structured high-dimensional data like
images, because individual input pixels lack interpretability and
Euclidean distance between images is not meaningful. We therefore
provide an 
extension
called \dbafull{}, which stands for DBA with
attributes. \dbafull{} measures distance in a latent representation
space $\latentspace$, which is learned by a VAE from the training
data~$D$. The user is further required to describe attributes that are
meaningful to them by providing additional data annotations. We use
these annotations to predict which attributes are present for any given
latent representation $\z \in \latentspace$. See
Figure~\ref{fig:attdba_scheme} for an overview of all the spaces and
mappings between them.

\paragraph{Learning the Latent Representation Space}

Variational autoencoders provide an unsupervised procedure to learn
(non-linear) mappings back and forth between inputs in $\inputspace$ and
latent representations in $\latentspace \subset \reals^l$. The
dimensionality $l$ of the latent space is taken to be much smaller than
the input dimension $d$, which forces a dimensionality reduction. VAEs
have the important property that the marginal distribution over the
latent space is approximately standard Gaussian, which means that
Euclidean distance is an appropriate metric in $\latentspace$. We do not
commit to any single choice of VAE, but allow the VAE to be customized
for the data under consideration. We further adjust the training
procedure of the VAE to favor preservation of class probabilities, which
can significantly improve the \emph{label stability}, i.e.\ $f(\x') =
f(\x)$, where $\x'$ is the reconstruction of $\x$ that is obtained by
mapping $\x$ to $\latentspace$ and back using the VAE. See
Appendix~\ref{app:dba} for details. Label stability for $\x_0$ is
crucial, because otherwise it is hopeless to use the VAE in any procedure
that tries to explain the decision boundary.

\paragraph{Predicting User-specified Attributes}

We assume the user provides annotations for $p$ attributes, where each
annotation $A_j = \big( (\x_1,y_1), \ldots, (\x_{n_j},y_{n_j})\big)$
consists of pairs of inputs $\x_i$ and binary labels $y_i \in \{-1,+1\}$
that indicate whether attribute $j$ is true or false. For example, an
attribute might indicate whether the person in an image is smiling or
not. The user may annotate (part of) the training data $D$, or provide
separate annotated data. For each annotation $A_j$, we first map the
inputs $\x_1,\ldots,\x_{n_j}$ to their latent representations
$\z_1,\ldots,\z_{n_j}$ using the VAE and then use $L_2$-penalized
logistic regression to train what we call an \emph{annotator} $a_j :
\latentspace \to [0,1]$, which predicts the probability that attribute
$j$ is true from the latent representation of an input. We denote the
coefficients of the logistic regression model for $a_j$ by $\esttheta_j
\in \reals^l$ and its intercept by $\hat{\theta}_{0,j}$. Thus, we can go
from any input $\x \in \inputspace$ to a latent representation $\z \in
\latentspace$ to a vector $\a = \big(a_1(\z),\ldots,a_p(\z)\big)$ of
attribute probabilities. We call the space that $\a$ lives in the
\emph{attribute space} $\attributespace \subset [0,1]^p$.

\begin{remark}
In special cases our approach might be simplified: if the dimensions in
the VAE latent space are directly interpretable for the user, then we
can identify each attribute with a dimension of the latent space, and no
annotations are needed.
\end{remark}

\paragraph{The \dbafull{} Method}

Our extended method \dbafull{} differs from \dbasimple{} as follows.
First, we map the point to be explained, $\x_0$, to its latent
representation $\z_0$. The detection step then runs in the latent space
$\latentspace$ instead of the input space $\inputspace$ and detects a
point $\zboundary$ on the decision boundary of $f$. Labels for any point
$\z \in \latentspace$ are obtained by mapping $\z \mapsto \x \mapsto
f(\x)$. Second, the simulation step also runs in $\latentspace$. Since
the coordinates of $\latentspace$ need not correspond directly to
interpretable features, we define the vertices in terms of the
user-specified attributes: $\v_{j,\pm 1} = \zboundary \pm \alpha
\esttheta_j$ for $j = 1,\ldots,p$. The interpretation is that
$\esttheta_j$ represents the direction that increases the probability of
attribute~$j$ being true. Finally, in the explanation step we fit the
linear surrogate model in the attribute space, by mapping each sample
point $(\z_i,y_i)$ to $(\a_i,y_i)$ using the annotators. To make the
coefficients of the surrogate model comparable between attributes, we
standardize all attributes $a_{i,j}$ based on their mean and standard
deviation in the sample $S$.

\section{Experiments}\label{sec:experiments}

To evaluate \dbafull{} we design two controlled experiments in which
interpretable attributes are available and we know how the class labels
are assigned based on these attributes. In the first experiment we
generate our own artificial data set of flower images (AIris). In the
second experiment we use the CelebA data set \cite{celeba}, for which
attribute annotations are available. In
Appendix~\ref{app:dbasimpleexperiments} we further include experiments
with the \dbasimple{} method on tabular data, including experiments on
standard UCI data sets in which \dbasimple{} consistently outperforms
both LIME and MAPLE in terms of its ability to approximate the nearest
decision boundary region.

\subsection{AIris: Artificial Iris Data}
\label{sec:AIris}

Inspired by Anderson's classical Iris flower data set
\cite{Anderson1935}, which was made famous by Fisher \cite{Fisher1936},
we have created an image generation program that generates $128 \times
128$ RGB images of flowers based on 5 continuous parameters that all
have ranges inside $[0,1]$. See Appendix~\ref{app:airis} for examples
and further details. The first four parameters control the shape of the
flower. They are petal length (PL), petal width (PW), sepal length (SL)
and sepal width (SW). The last parameter, color (C), is a mixing
parameter that interpolates between red and magenta. We have used this
program to generate a training set $D$ of $4000$ flowers, as well as a
test set with $2000$ images. Each image was generated independently by
sampling the five parameters from the uniform distribution over their
range. We assign a flower to class~A if
\begin{equation}
     \begin{aligned}
       0.33\PL + 0.33\PW + 0.33\C &< 0.5 \quad \text{and}\\
       0.33\PL + 0.33\PW + 0.33\SL &> 0.4,
    \end{aligned}
\label{eqn:hyperplanes}     
\end{equation}
and to class~B otherwise. This assignment defines a non-linear ground
truth consisting of two hyperplanes that are defined in terms of latent
parameters which are not available to the classifier~$f$. The
hyperplanes were chosen to achieve approximately balanced classes. We
further annotate the training data with binarized versions of the
parameters: for each parameter $\PL,\PW,\ldots,\C$ we define an attribute
that measures whether the parameter value is large or not. We set this
attribute to $+1$ if the parameter exceeds the midpoint of its range,
and to $-1$ otherwise. Thus the parameters are also not directly
available to the explanation methods. The appeal of the AIris data is
that it is sufficiently simple for a standard convolutional VAE to
learn good latent parameters, but sufficiently difficult to illustrate
the differences between existing explanation methods.

\subsubsection{Experiments and Results}

We train a 5-layer CNN $c : \inputspace \to [0,1]$ on
the training data to learn the probability of class~A. The
corresponding binary classifications are: $f(\x) = +1$ if $c(\x) > 0.5$
and $f(\x) = -1$ otherwise. See Appendix~\ref{app:airis} for
details. The CNN achieves 99.33\% accuracy on the training data and
98.75\% accuracy on the test set, which is sufficiently high that, on
images from the same source, its decision boundary must be very similar
to the ground truth specified by the two hyperplanes. As described in
Appendix~\ref{app:airis}, we further train a convolutional VAE on the
training data. The VAE achieves 90\% label stability on the test set (up
from 74\% without our adjustment to favor preservation of class
probabilities). We compare \dbafull{} to the CEM-MAF pertinent negative
method \cite{CEM-MAF}. We also want to compare to LIME, but since LIME
\cite{LIME} and ALIME \cite{ALIME} do not have the option to learn
user-specified attributes from annotations, we instead compare to a
hybrid method \limefull{}, which is a variant of \dbafull{} in which we
have replaced the detection and simulation steps by LIME. We also
include a non-local baseline that we call the \globalsurrogate{} method.
\globalsurrogate{} is given an advantage because it gets as features the
original parameters ($\PL, \PW, \ldots, \C$) used to generate the
images, but it is also at a disadvantage because it fits a global linear
model in terms of these parameters, so it cannot exploit locality. We
run \dbafull{} with $k = 1000,m = 500$ and $r$ tuned automatically from
grid $\rgrid = \{0.1,0.2,\ldots,0.9,1,1.5,2,\ldots,9.5,10\}$. For
details about the other methods, see Appendix~\ref{app:airis}.
\dbafull{}, \limefull{} and \globalsurrogate{} all output a coefficient
vector $\estbeta \in \reals^p$ that expresses the relative importance of
the attributes. This corresponds to a direction $\latentdirection =
\sum_{j=1}^p \hat{\beta}_j \esttheta_j$ in the latent space
$\latentspace$, where $\esttheta_j \in \reals^l$ are the coefficients of
the annotator for attribute~$j$ (see Section~\ref{sec:dbafull}). CEM-MAF
outputs a contrastive example $\xcontrast$, but, as described in
Appendix~\ref{app:airis}, it may also be used to obtain a direction
$\latentdirection$ in the latent space and a vector of coefficients
$\estbeta$ for the attributes.

\begin{figure}[htbp]
  \centering
  \includegraphics[width=0.28\textwidth]{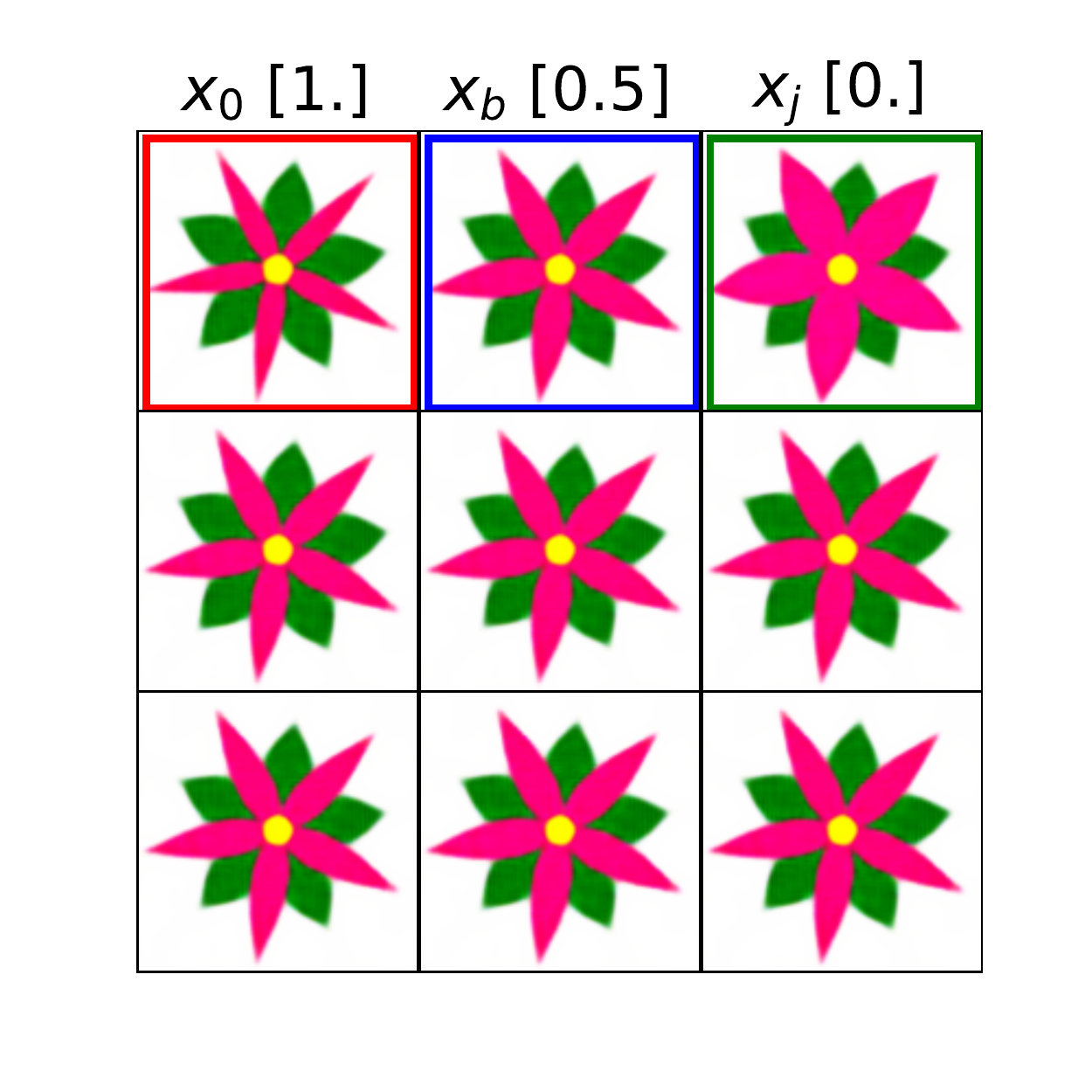}
  \caption{Illustration of \dbafull{} detection and sampling steps}
  \label{fig:samples_aires}
\end{figure}

\paragraph{A First Impression}

Before reporting aggregate statistics for multiple explanations, we
first illustrate our results on a single representative input image
$\x_0$.
Two other inputs illustrating best and worst performance for
\dbafull{} are shown in Appendix~\ref{app:airis}.
Figure~\ref{fig:samples_aires} shows the input $\x_0$ in a red frame. It
further shows the closest decision boundary point $\xboundary$ in a blue
frame, and in a green frame there is the corresponding bisected point
$\x_j$ from the detection step of the algorithm. The other images are
6~random samples from $S$ that were generated in the simulation step.
Appendix~\ref{app:airis} shows the principal component analysis (PCA)
projection of $S$ onto two dimensions. It can be seen there that the two
classes occur in roughly equal proportions, and can be separated quite
well with a linear decision boundary. The corresponding linear local
surrogate model $g$ is indeed highly faithful to $f$: its fidelity is
$99.7\%$ on $S$.

\begin{figure}[htbp]
    \centering
    \begin{subfigure}[t]{0.55\textwidth}
        \centering
        \includegraphics[width=0.65\textwidth]{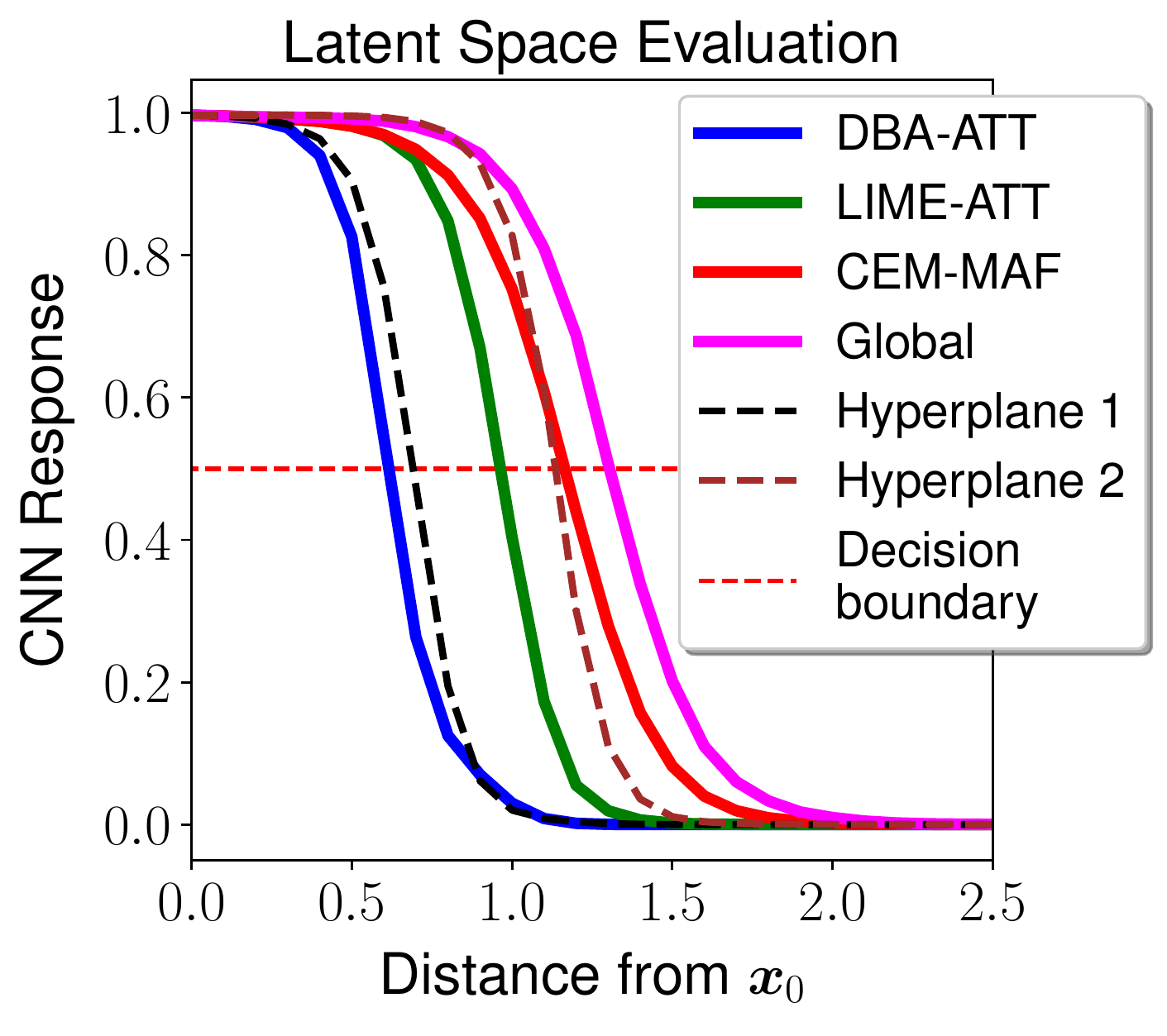}%
        \caption{CNN probabilities moving in direction
        $\latentdirection/\|\latentdirection\|$ from $\x_0$}
        \label{fig:161_latentspace_airis}
    \end{subfigure}%
    \begin{subfigure}[t]{0.45\textwidth}
        \centering
        \includegraphics[width=0.94\textwidth]{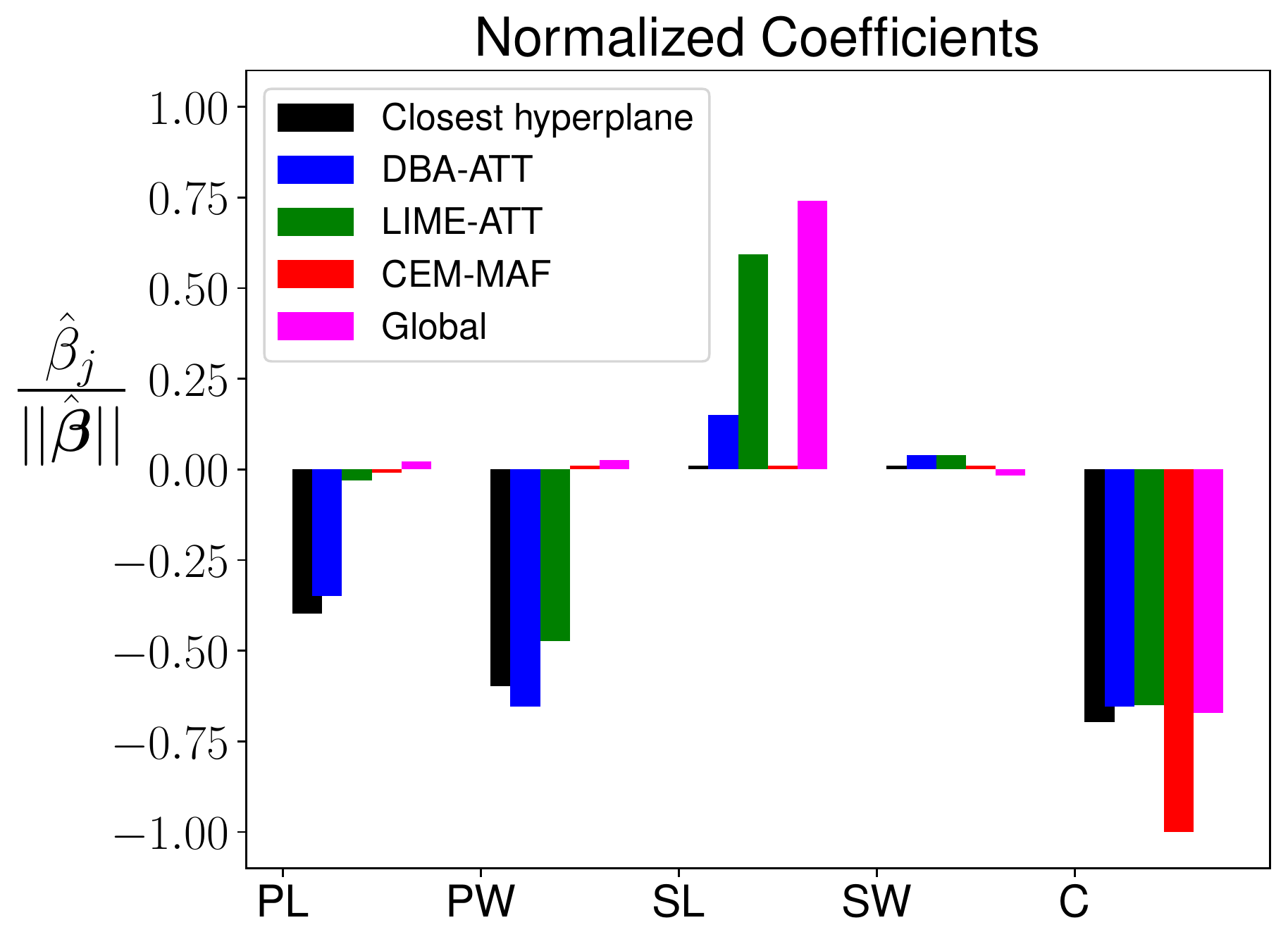}
        \caption{Explanation coefficients $\estbeta$}
        \label{fig:161_coeffs_airis}
    \end{subfigure}
    \caption{Evaluation of explanation directions on the AIris
    test image from Figure~\ref{fig:samples_aires}}
    \label{fig:comparison_aIris}
\end{figure}

Figure~\ref{fig:comparison_aIris} compares the quality of explanations
on a single test image. We see in Figure~\ref{fig:161_latentspace_airis}
that \dbafull{} has found a direction that gets to the decision boundary
much faster than the other methods. Both figures in
Figure~\ref{fig:comparison_aIris} also show that the \dbafull{}
direction is similar to the direction of the closest
hyperplane\footnote{\label{foot:hyperplane}For proper comparison, we
standardize each coefficient of the hyperplane by multiplying it by the
standard deviation of its corresponding parameter. See
Appendix~\ref{app:airis}.}, whereas the directions for the other methods
are different. We proceed to show that the behaviors observed in these
examples in fact hold generally.

\begin{table*}
  \caption{AIris evaluation statistics, averaged over $50$ test images}
  \label{tab:global_measures_airis}
  \centering
  \footnotesize
  \begin{tabular}{llllllll}
    \toprule
                    & DBA      & LIME           & Global    & Class           & Latent       & Cosine         & Cosine         \\
                    & Fidelity & $R^2$-Fidelity & Fidelity  & Balance         & Distance     & Similary-      & Similarity+   \\
  \midrule
 \dbafull{}         &  97.0\%  & -              &    -      & \textbf{50.2\%} & \textbf{0.6} & 0.858 & \textbf{0.925} \\
 \limefull{}        &        - & 21.7\%         &    -      & 47.7\%          & 1.0          &  0.570        & 0.719 \\
 CEM-MAF            &     -    &   -            &    -      &    -            & 1.0          &  0.458        & 0.640\\
 \globalsurrogate{} & -        & -              &  73.6\%   & 47.4\%          & 1.8          & 0.403          & 0.454\\
    \bottomrule
  \end{tabular}
\end{table*}

\paragraph{Aggregate Evaluation}

All methods are evaluated on the same set of $50$ randomly selected
images $\x_0$ from the test data on which the VAE achieves label
stability. (See Section~\ref{sec:dbafull} for a discussion on label
stability.) In Table~\ref{tab:global_measures_airis} we report the means
of the following statistics. First, fidelity measures how well the
surrogate models approximate the original classifier $f$. Unfortunately,
there is no single measure of fidelity, because the methods generate
different samples in different regions of the input space, and it does
not make sense to evaluate how well one method fits a sample generated
by another method. For \dbafull{} we therefore measure ``DBA Fidelity''
by the classification accuracy of the surrogate in predicting the labels
of $f$ on the local sample $S$ generated by \dbafull{}; for ``LIME
Fidelity'' we use the local Weighted Regression $R^2$ measure proposed
by the LIME authors on the \limefull{} sample \cite{ALIME}; and for
\globalsurrogate{} we measure ``Global Fidelity'' by its classification
accuracy on all the training data. Since CEM-MAF does not directly
produce a surrogate model, we do not measure its fidelity. Second, we
measure the class balance of the samples produced by \dbafull{} and
\limefull{} by the percentage of sample points labeled as $+1$ by $f$.
For \globalsurrogate{} we report the class balance of all the training
data. Third, we measure the ``Latent Distance'' from $\z_0$ to the
decision boundary in the latent space along the direction
$\latentdirection$. Finally, we include two measures that capture the
cosine similarity between $\estbeta$ and the coefficients of the true
hyperplanes in \eqref{eqn:hyperplanes}.\cref{foot:hyperplane} The
difference is that ``Cosine Similarity-'' measures similarity with the
closest of the two hyperplanes and ``Cosine Similarity+'' measures the
maximum of the similarities with the two hyperplanes. 

\paragraph{Aggregate Results}

For \dbafull{} fidelity is high, while fidelity is low for \limefull{}:
its local surrogate model is only able to explain $21.7\%$ of the
variance in the class probabilities of the CNN $c$. We further see that
class balance is good (close to $50\%$) for all methods. From Latent
Distance we see that \dbafull{} is much better at identifying the
nearest decision boundary direction than \limefull{} and CEM-MAF, which
perform equally well. \globalsurrogate{} is worse still, showing that
local explanations are much more informative than a single global
explanation. Finally, we see from Cosine Similarity+ that \dbafull{}
recovers the coefficients of one of the hyperplanes extremely well: the
similarity is close to its maximum value $1$. The other methods are
significantly worse. The gap between the two Cosine Similarities for all
methods indicates that the methods do not necessarily focus on the
closest hyperplane. This can be explained by the fact that, for a subset
of data points from class B, the closest hyperplane does not switch the
class.

\subsection{CelebA: Annotated Celebrity Image Data}
\label{sec:celeba}

\begin{table}[htbp]
    \centering
    \caption{CelebA evaluation statistics, averaged over $30$ test
  images\label{tab:global_measures_celeba}}
      \begin{tabular}{lllll}
     \toprule
     & DBA      & LIME           & Class   & Latent   \\
     & Fidelity & $R^2$-Fidelity & Balance & Distance \\
   \midrule
    \dbafull{}  &  97.3\% & -   &  \textbf{50.1\%} & \textbf{2.0} \\
    \limefull{} &   -  & 32.7\% &  47.2\%     & 2.1 \\
    CEM-MAF     &   -  & -   &             - & 2.7 \\
     \bottomrule
   \end{tabular}
\end{table}

\begin{figure}[htbp]
  \centering
  \includegraphics[width=0.5\textwidth]{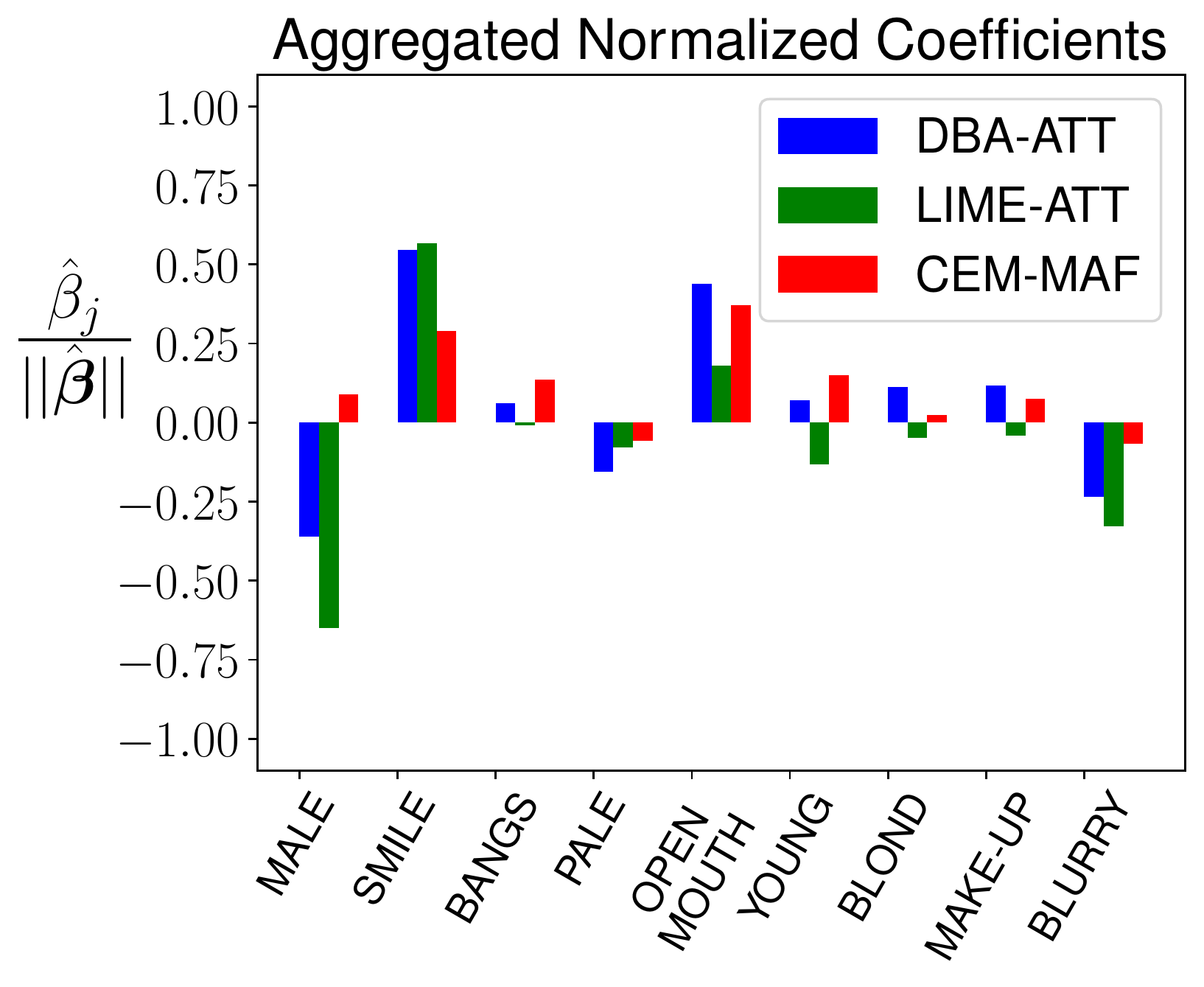}
  \caption{Mean explanation coefficients for CelebA} \label{fig:avgcoeffCelebA}
\end{figure}

\begin{figure}[htbp]
  \begin{subfigure}[t]{0.38\textwidth}
     \centering
     \includegraphics[width=0.76\textwidth]{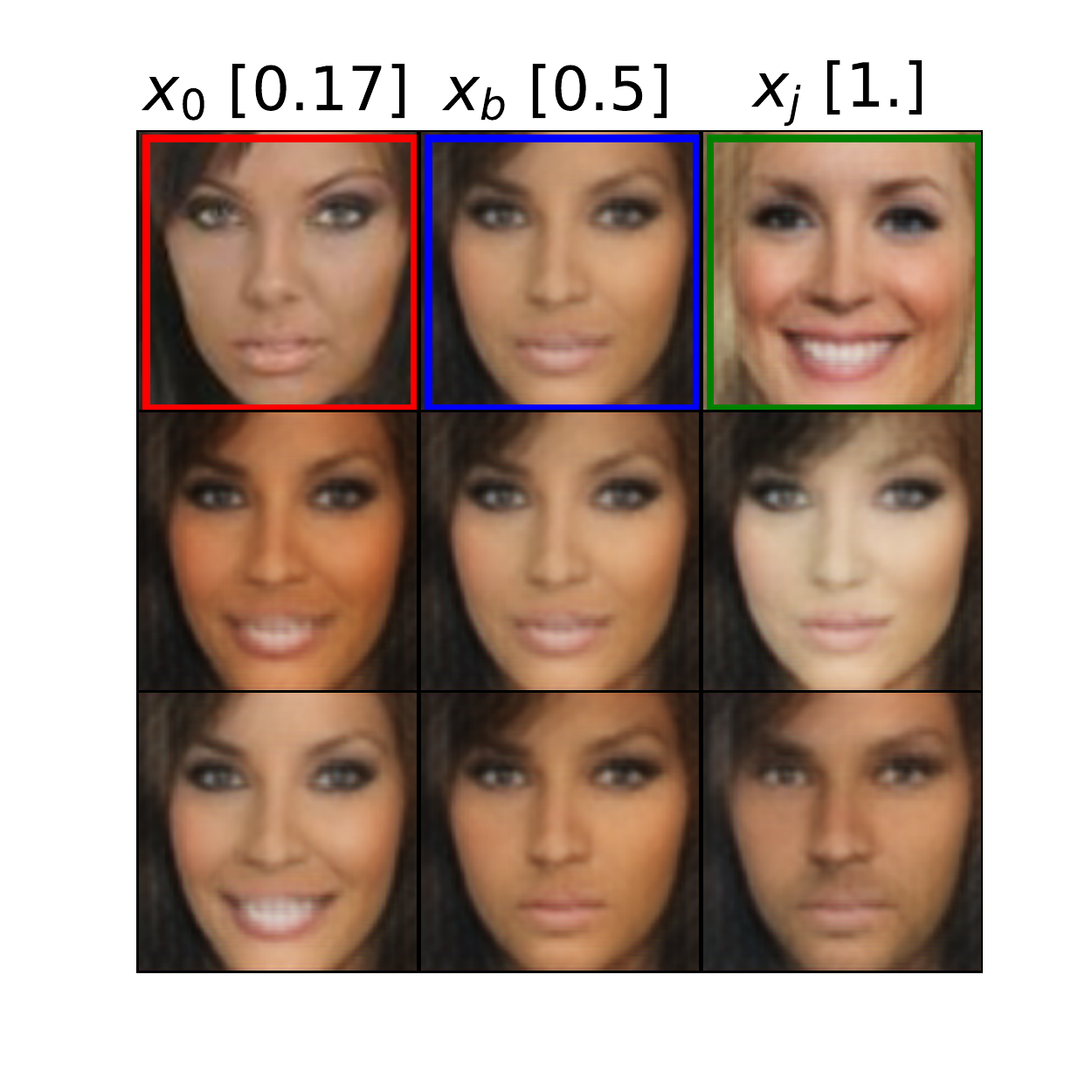}
     \caption{Illustration of detection and sampling steps for
     \dbafull{}}\label{fig:24samples}
  \end{subfigure}%
  \begin{subfigure}[t]{0.32\textwidth}
    \centering
    \includegraphics[width=\textwidth]{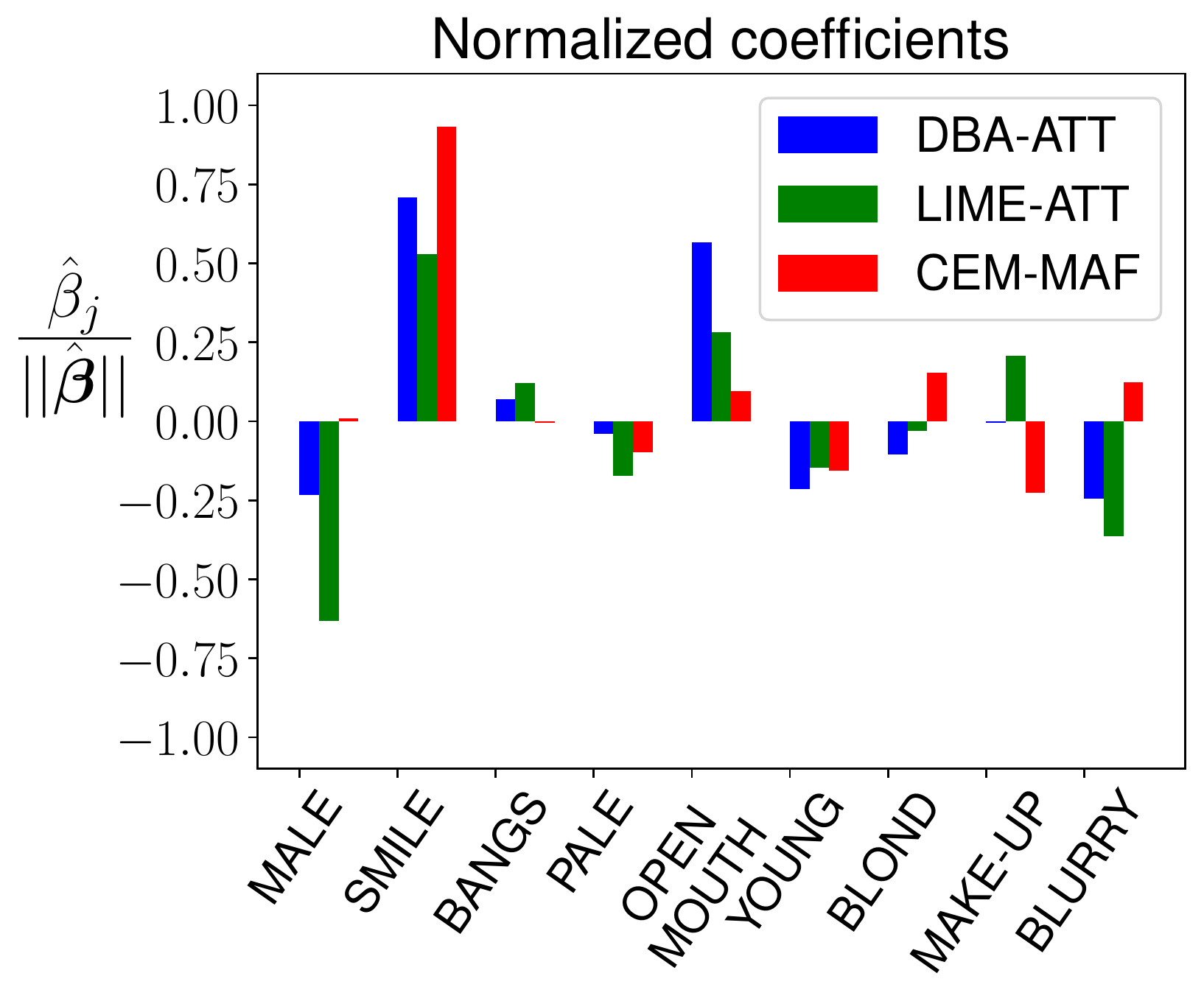}
    \caption{Explanation coefficients}
  \end{subfigure}%
  \begin{subfigure}[t]{0.30\textwidth}
    \centering
    \includegraphics[width=0.94\textwidth]{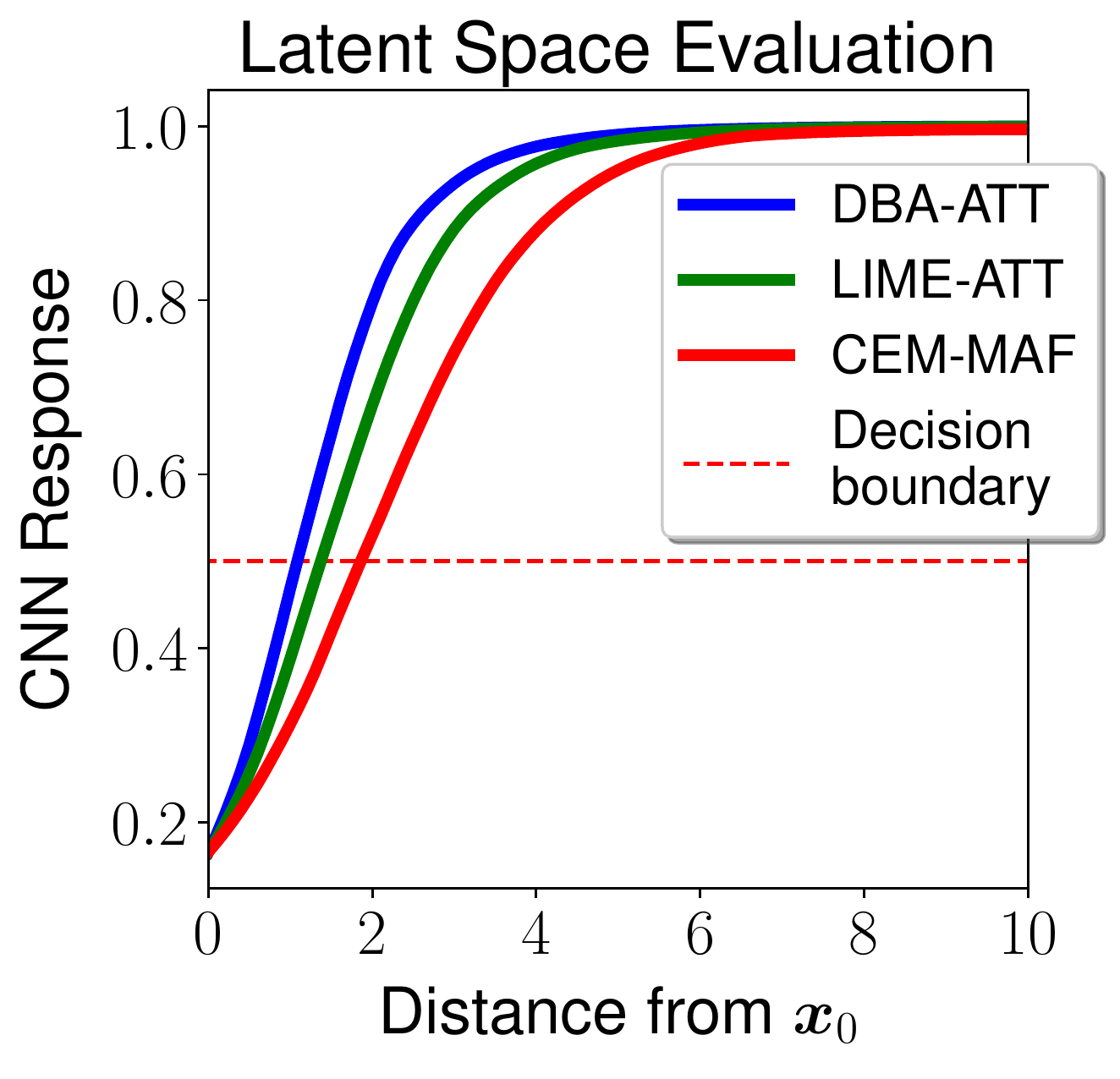}
    \caption{CNN probabilities as we move in direction
    $\latentdirection/\|\latentdirection\|$ from $\x_0$}
    \label{fig:24path}
  \end{subfigure}%
  \caption{Explanations of the single image shown in the top-left corner
  of Figure~\ref{fig:24samples}}\label{fig:example24}
\end{figure}

CelebA is a large image data set of 202\,599 celebrity faces, annotated
with 40 binary attributes~\cite{celeba}. We restrict attention to 9 attributes: MALE, SMILE, BANGS, PALE, OPEN MOUTH, YOUNG, BLOND, MAKE-UP
and BLURRY, and define two classes: A) SMILE and not MALE $\approx$
smiling females vs B) other. Class A was chosen because it is one of the
few combinations of two attributes that gives approximately balanced
classes. We split the data into $n=162\,079$ training images (80\%) and
40\,520 test images (20\%). A CNN is trained with 93.2\% training
accuracy and 91.5\% testing accuracy. We further train a VAE based on
\citet{DFCVAE}, which gives 91\% label stability. For the
explanation methods, we only use the annotations from 9\,000 randomly
selected images from the training data, and we explain 30 random images
from the test set on which the VAE achieves label stability using
\dbafull{}, \limefull{} and CEM-MAF. See
Appendix~\ref{app:CelebAdetails} for further details.

\subsubsection{Experiments and Results}

In Table~\ref{tab:global_measures_celeba} we report the same aggregate
statistics as for the AIris data, except for the ones that require
knowledge of the true data generating mechanism. We see that \dbafull{}
and \limefull{} are approximately equally good in terms of distance to
the decision boundary, which suggests that for this data the local CNN
probabilities increase more or less orthogonally from the decision
boundary. In this case, CEM-MAF is significantly worse at indicating the
direction of the nearest decision boundary. A possible explanation why
the low $R^2$-fidelity of \limefull{} does not prevent it from achieving
small Latent Distance is that its surrogate model might approximate the
decision boundary reasonably well even if it cannot approximate the CNN
probabilities very precisely.

In Figure~\ref{fig:avgcoeffCelebA} we further show the mean
coefficient values for the three methods. Both \dbafull{} and \limefull{} detect
the importance of MALE and SMILE, and further show that the CNN is
improperly sensitive to OPEN MOUTH and BLURRY. The effect of blurriness
is particularly interesting because it cannot be expressed by
highlighting the most important pixels, which is a common approach in
explainability \cite{SelvarajuEtAl2017GradCam,LIME}. CEM-MAF
disqualifies itself because it gives the wrong dependence on the MALE
attribute.

Zooming in on a single image, Figure~\ref{fig:example24} shows a very
similar pattern to the aggregate explanations, with perhaps increased
importance of SMILE and OPEN MOUTH.

\section{Conclusion}
\label{sec:discussion}

We have introduced the AIris data set as a new high-dimensional
benchmark for local explanations that focus on the decision boundary.
The importance of AIris is that the ground truth is known, so we can
directly evaluate how good different methods are at recovering it. Our
general non-tabular method \dbafull{} produces explanations that better
identify the nearest decision boundary region than both \limefull{} and
CEM-MAF for both the AIris and the CelebA data.
Appendix~\ref{app:tabularUCI} further shows that \dbasimple{}, the basic
version of our method, consistently outperforms LIME and MAPLE on
tabular data in terms of distance to the decision boundary.

Our DBA procedures fit a local linear surrogate model to the nearest
decision boundary point. As illustrated in Figure~\ref{fig:dba_toy}, the
coefficients $\estbeta$ of this model may be interpreted as a vector
pointing towards the nearest boundary point. For users who would prefer
a contrastive explanation in terms of data point $\xcontrast$ sampled
from the opposite class compared to $\x_0$, it would be possible to
provide such a contrastive explanation by moving along the vector
$\estbeta$ until we cross the decision boundary, and then sampling
$\xcontrast$ there.

Our approach requires a successfully trained VAE and sufficiently
expressive attributes: a linear model in terms of the attributes, which
corresponds to a linear model in a subspace of the latent space of the
VAE, must be able to locally approximate the decision boundary of the
classifier $f$ with high fidelity. A less obvious point of attention is
that user annotations are not always univocal: if a user annotates a
data set in which all men have short hair and all women have long hair,
then the corresponding attribute MALE will not just correspond to gender
but also to hair length. This is the problem of \emph{entangled} latent
representations, which recent work on VAEs is starting to address
\cite{pmlr-v97-mathieu19a,NIPS2019_9603,locatello2019disentangling}.

\small
\bibliographystyle{abbrvnat}
\bibliography{references.bib}
\normalsize


\appendix

\section{DBA}
\label{app:dba}

This first section of the Appendix contains additional material
related to Section~\ref{sec:dba}. We first present experiments
illustrating the behavior of \dbasimple{}, and then we provide
additional details that were omitted from the description of \dbafull{}
in Section~\ref{sec:dbafull}.

\subsection{Experiments with \dbasimple{}}
\label{app:dbasimpleexperiments}

In this section we provide additional experiments on tabular data for
the simplest version of our method, \dbasimple{}. We compare with LIME
and report similar aggregate statistics as for the two main experiments
in the paper. In the first experiment we illustrate a 2D toy case (the
Moons data set), in the second experiment we consider a simplified
tabular version of the AIris experiment from Section~\ref{sec:AIris},
and in the third experiment we provide a real-world example where
\dbasimple{} and LIME draw opposite conclusions about the importance of
a particular feature. Finally, we also compare \dbasimple{} to both LIME
and MAPLE on four standard tabular data sets from the UCI repository.

\subsubsection{Moons Data: a 2D Toy Example}
\label{app:moons}

The Moons distribution is a standard toy example that generates two
moon-shaped classes in two dimensions that are not linearly separable.
We sample 1000 points from this distribution with noise parameter 0.15,
as implemented in Scikit-learn~\cite{scikitlearn}. We split the data set
in $n=600$ training samples and $400$ testing examples, and train a
support vector machine (SVM) with radial basis function kernel
$\exp(-\tfrac{1}{2}\|\x - \x'\|^2)$ and regularization parameter $C=1.0$
on the standardized training data. \dbasimple{} can work directly with
the decision boundary for the SVM, but LIME requires probabilities to
produce its explanations, which are not directly available from an SVM.
We therefore map the SVM margins to a probability by Platt scaling
\cite{Platt99} based on 5-fold cross-validation on the training set, as
implemented in Scikit-learn. It is known that Platt scaling can change
the decision boundary, but in the present case agreement between the
classifications based on the probabilities and those of the original SVM
is 99.8\% on the test set, so the change is minor. The resulting
classifier achieves 98\% accuracy on the test set.

\begin{figure}[h!]
  \centering
  \begin{subfigure}[t]{0.30\textwidth}
     \centering
     \includegraphics[width=\textwidth]{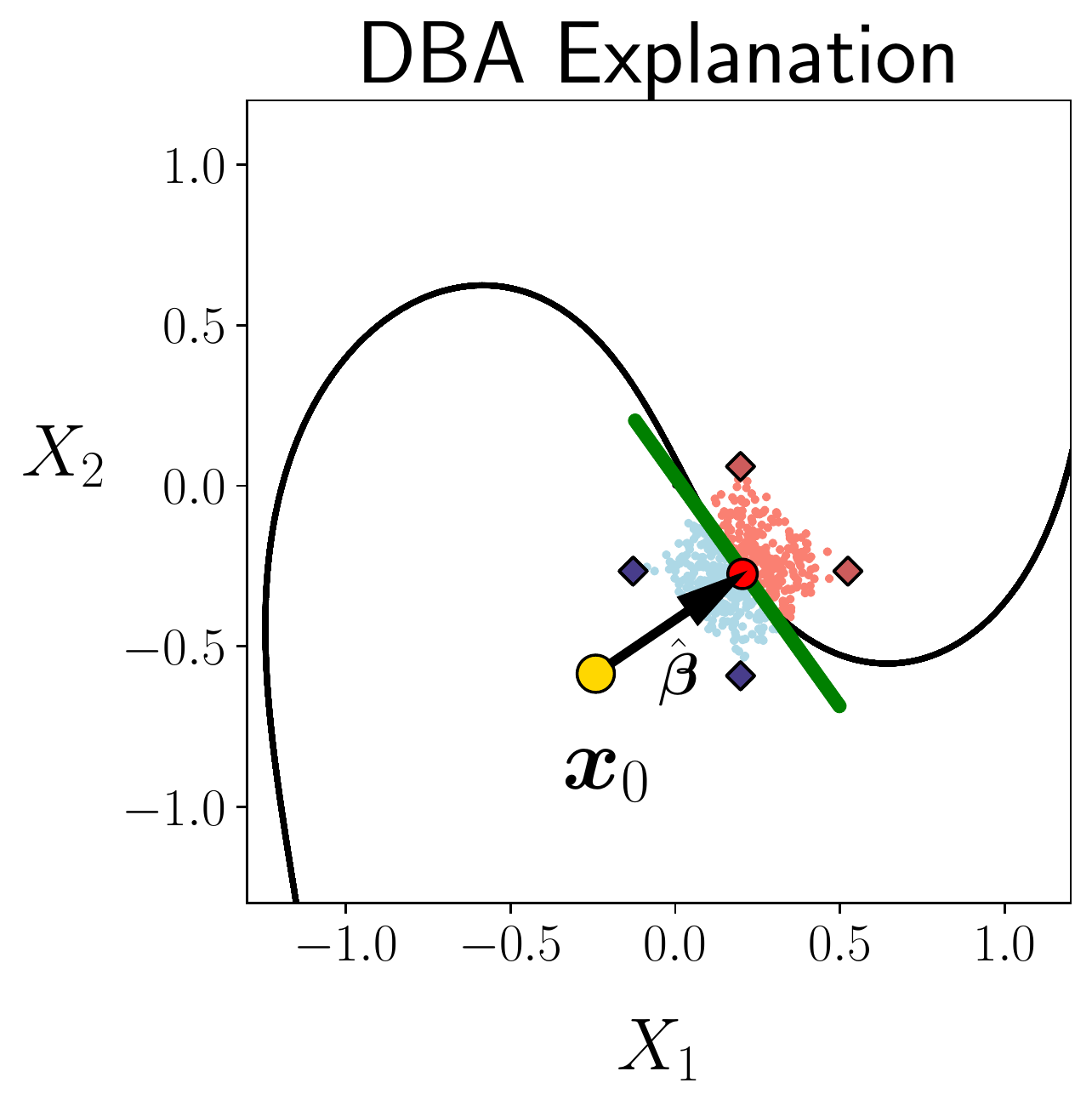}
     \caption{\dbasimple{}}\label{fig:solution_moons_dba}
  \end{subfigure}%
  \hfill
  \begin{subfigure}[t]{0.30\textwidth}
    \centering
    \includegraphics[width=\textwidth]{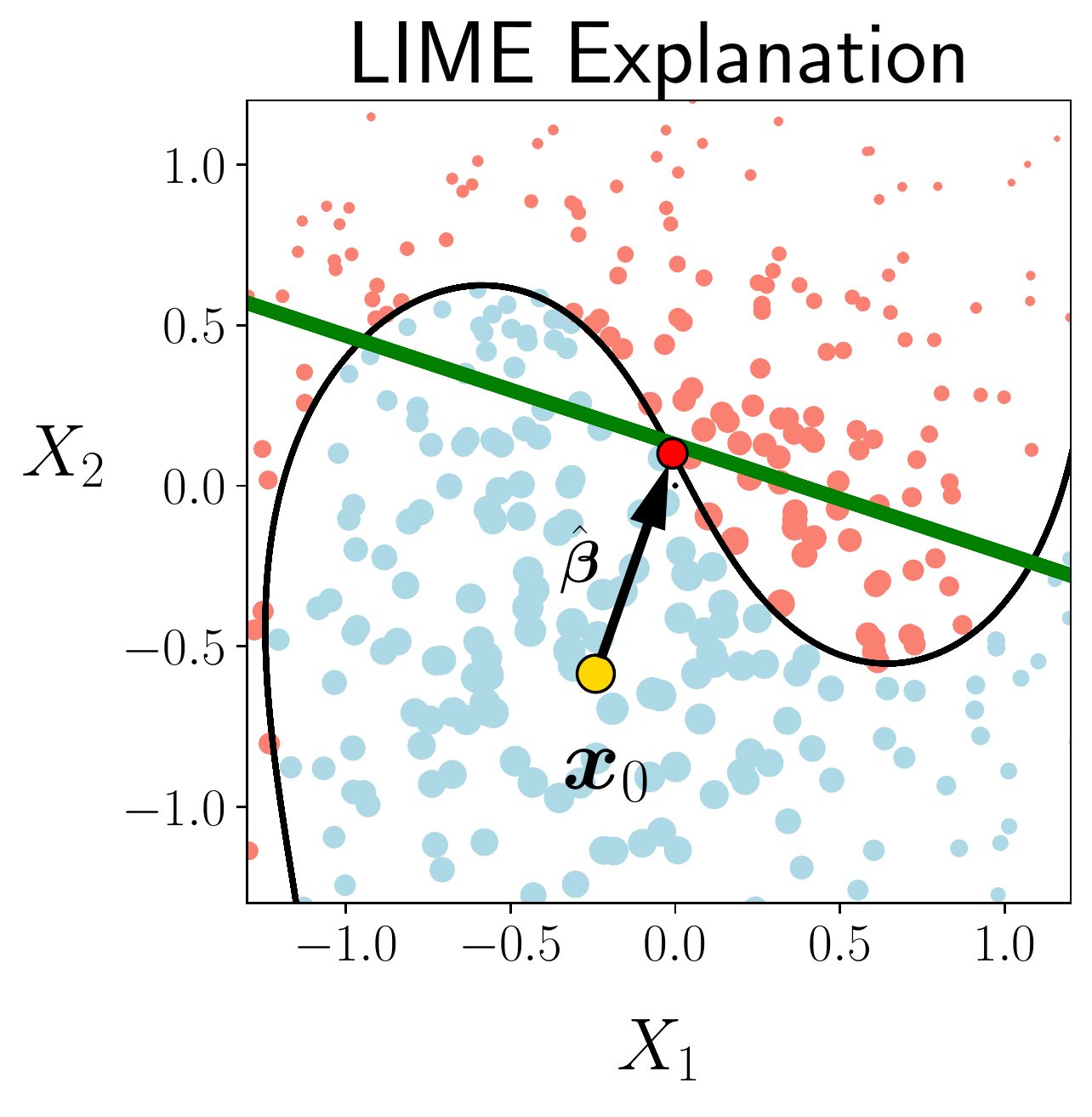}
    \caption{LIME outputs an average of the decision boundary instead of
    a single part of it. (The size of points reflects their weight when fitting
    LIME's local surrogate model.)}
  \end{subfigure}%
  \hfill
  \begin{subfigure}[t]{0.30\textwidth}
    \centering
    \includegraphics[width=\textwidth]{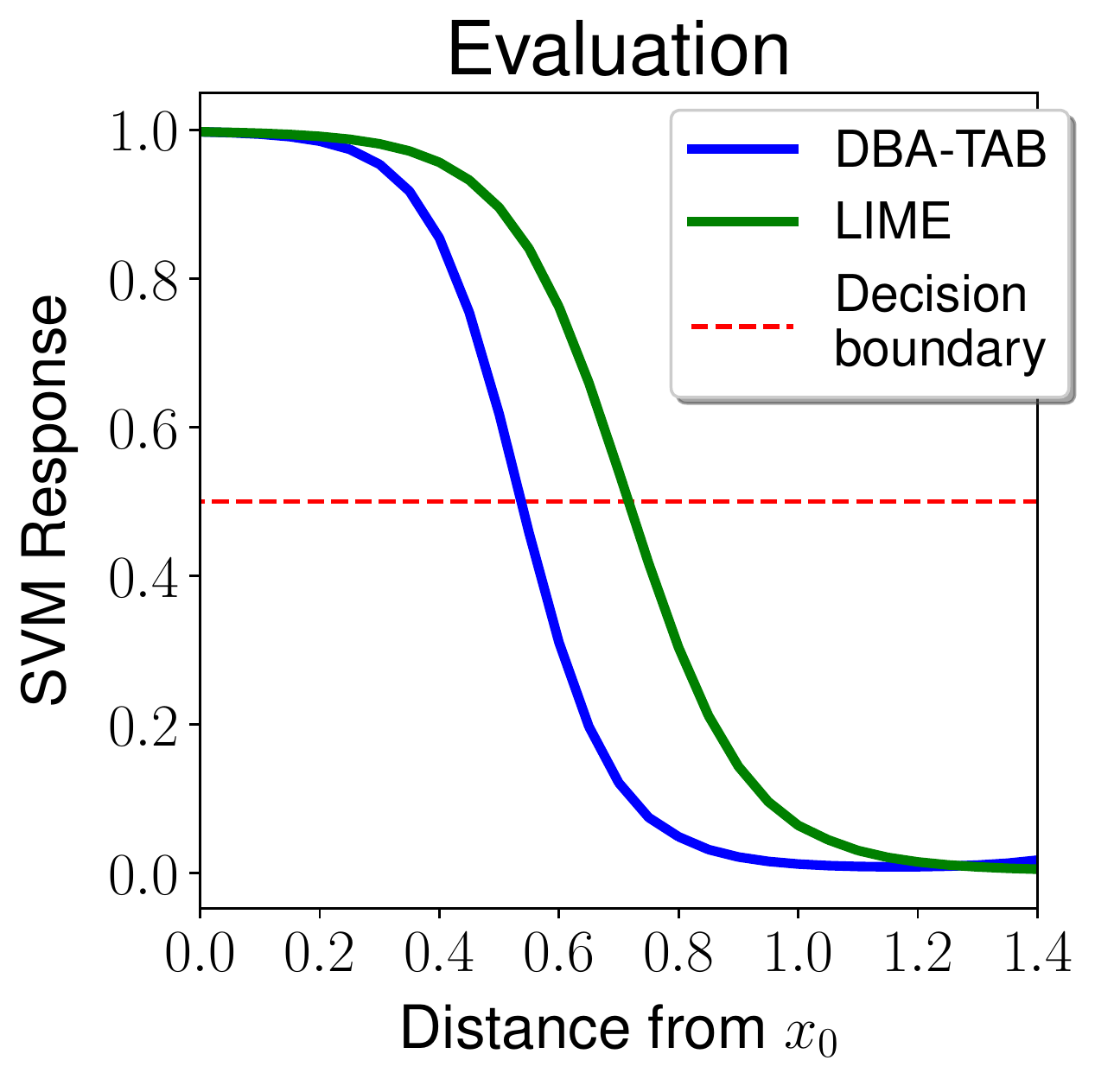}
    \caption{SVM probabilities as we move in direction
    $\estbeta/\|\estbeta\|$ from $\x_0$}
    \label{fig:moonslatent}
  \end{subfigure}%
  \caption{Comparison of \dbasimple{} and LIME solutions for the Moons
  data}
  \label{fig:dba_lime_comparison_moons}
\end{figure}

\paragraph{Experiments and Results}

For \dbasimple{} we use $k = m = 500$ and we take the grid of possible
values for $r$ to be $\rgrid = \{0.2,0.3,\ldots,1.5,2,2.5,\ldots,5\}$.
This $\rgrid$ is different from Section~\ref{sec:experiments} to make
sure all $r \in \rgrid$ are large enough to be visualized easily in
Figure~\ref{fig:solution_moons_dba}. The LIME paper \cite{LIME} does not
specify how to sample for tabular data with continuous features, so we
follow the approach in the LIME software~\cite{LIMESoftware2020} and
sample from a multivariate normal distribution
$\normaldist(\empmean,\empSigma)$, where $\empmean$ is the empirical
mean of the train set and $\empSigma = \diag(s_1^2,s_2^2)$ is a diagonal
matrix based on the empirical variances $s_1^2$ and $s_2^2$ of $X_1$ and
$X_2$ in the training data. For LIME we sample $m=500$ points. We
further use the standard weights $\exp(-\|\x - \x_0\|^2/\sigma^2)$ for
the default choice $\sigma = 0.75\sqrt{d}$ for dimension $d=2$. The
standard LIME implementation uses the LASSO to preselect a small subset
of the features before fitting its surrogate model, but since this is
very low-dimensional toy data, we omit this preselection step.

We first illustrate the difference between \dbasimple{} and LIME by
explaining a single data point, as shown in
Figure~\ref{fig:dba_lime_comparison_moons}. We see that \dbasimple{}
samples in a region centered on the decision boundary, whereas LIME
generates samples centered on $\x_0$ and then weighs them based on their
distance from $\x_0$. The local surrogate model for \dbasimple{} fits
the part of the decision boundary that is closest to $\x_0$, whereas the
surrogate model for LIME produces an average over a larger region of the
decision boundary that is not sufficiently local to be approximated well
by a linear model. It may be possible in this example to get a better
approximation of the decision boundary by decreasing $\sigma$ in an ad
hoc manner depending on the distance of $\x_0$ to the decision boundary,
but this would go against the spirit of LIME and it does not seem likely
that such tuning would be possible in general. Another effect, which is
not very visible in the figure, is that LIME is biased to avoid mistakes
on the class of $\x_0$. This happens because samples from the same class
as $\x_0$ tend to be closer to $\x_0$ than samples from the other class,
and therefore receive a higher weight. As Figure~\ref{fig:moonslatent}
shows, the \dbasimple{} explanation also corresponds to a direction that
crosses the decision boundary faster than the explanation for LIME.

\begin{table}[htb]
  \caption{Moons evaluation statistics, averaged over the whole training set}\label{tab:global_measures_moons}
  \centering
  \footnotesize
   \begin{tabular}{lllll}
     \toprule
     & DBA      & LIME           & Class   & Decision Boundary  \\
     & Fidelity & $R^2$-Fidelity & Balance & Distance \\
   \midrule
    \dbasimple{}  &  92.9\% & -   &  49.5\% & \textbf{0.67} \\
    LIME &   -  & 33.3\% &  49.4\%     & 0.81 \\
    
     \bottomrule
   \end{tabular}%
\end{table}

The general pattern that \dbasimple{} points more directly at the
decision boundary is confirmed by Table~\ref{tab:global_measures_moons},
which shows aggregate statistics when explaining all points from the
training data. We see that the mean fidelity for \dbasimple{} is still
high, especially compared to LIME, but a little lower than in the AIris
and CelebA experiments from Section~\ref{sec:experiments}, which
suggests that the decision boundary of the SVM in the current experiment
is locally less linear. Class balance of the generated samples of both
methods is close to 50\%. We note however that this does not mean that
the LIME sample equally represents both classes, because samples from
the class of $\x_0$ are generally closer and therefore receive a higher
weight than samples from the other class.

\subsubsection{A Tabular Simplification of AIris}
\label{app:tabairis}

The AIris experiment from Section~\ref{sec:AIris} jointly evaluates all
components of \dbafull{}, which includes the CNN that is being
explained, the VAE and the annotators. Here we provide a greatly
simplified tabular version of the experiment, which strips away as many of the
complications as possible and only evaluates \dbasimple{}. Since the
results are similar, we conclude that the simplified experiment captures
the essential parts of what is going on.

In our tabular simplification, no images are generated. Instead, we
directly observe the five parameters: the observations $\x = (\PL,
\PW, \SL, \SW, \C)$ come from the same uniform distribution as in
Section~\ref{sec:AIris} and are assigned to classes $A$ and $B$ according
to the same rule from \eqref{eqn:hyperplanes}. In fact, we do not
resample, and use exactly the same latent parameters that were used for
the sample from Section~\ref{sec:AIris} to obtain a train set of size
$n=4000$ and a test set of size $2000$.

\begin{figure*}[htb]
    \centering
    \hfill
    \begin{subfigure}[t]{0.45\textwidth}
        \centering
        \includegraphics[width=0.85\textwidth]{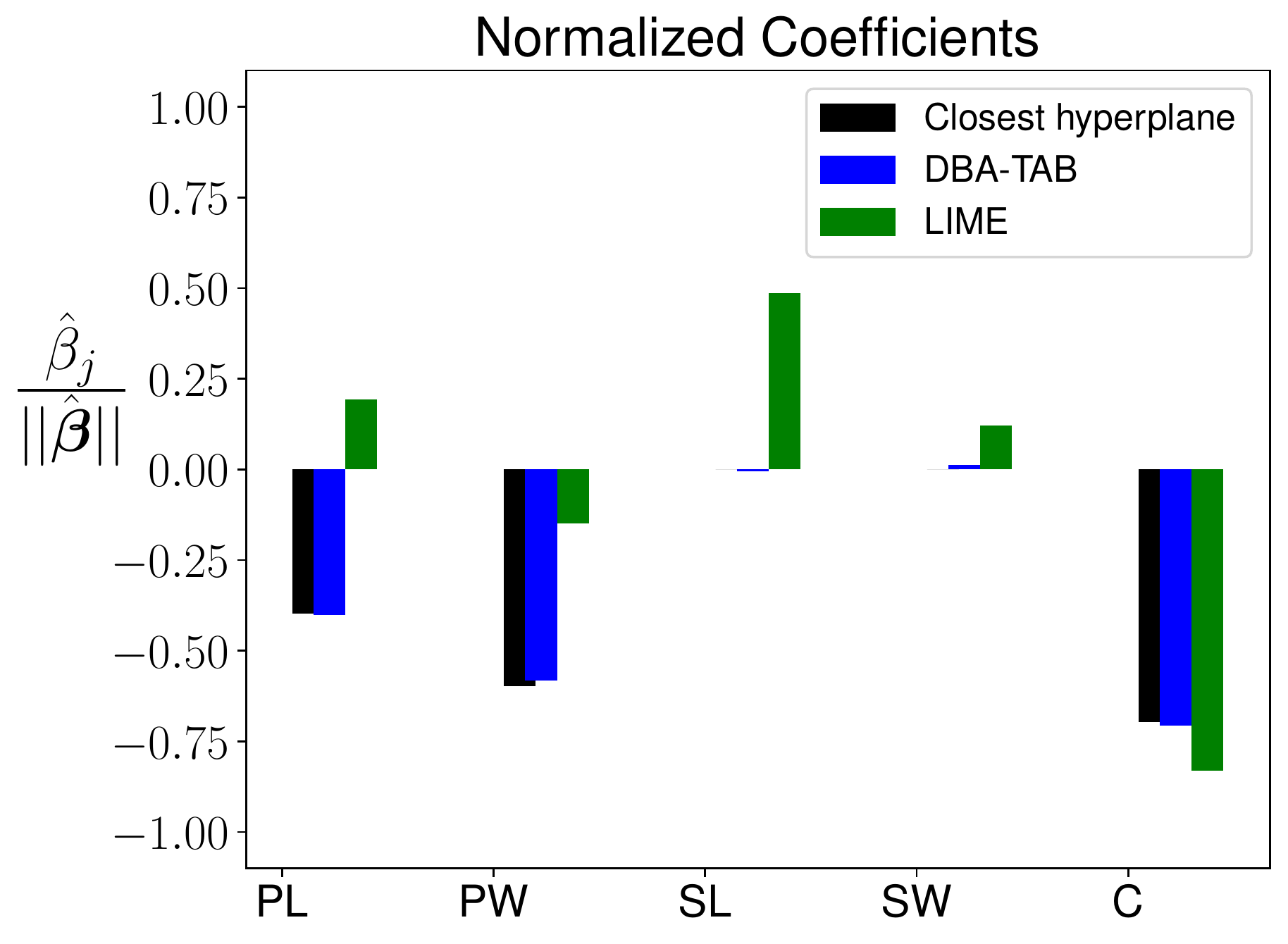}%
        \caption{Explanations for \dbasimple and LIME in simplified experiment from
        Appendix~\ref{app:tabairis}}
        \label{fig:161_latentspace_airis_true}
    \end{subfigure}%
    \hfill
    \begin{subfigure}[t]{0.45\textwidth}
        \centering
        \includegraphics[width=0.85\textwidth]{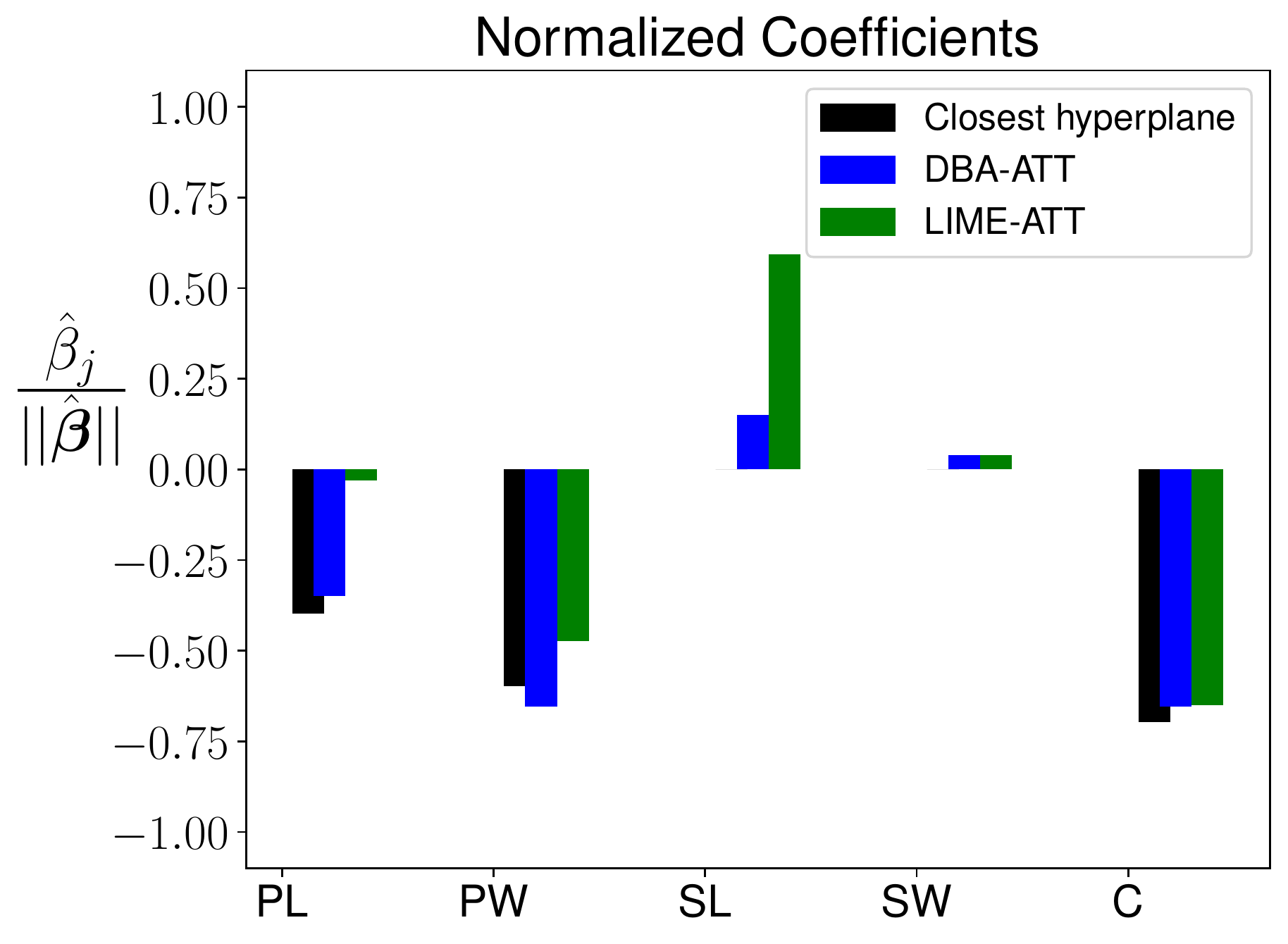}
        \caption{Explanations for \dbafull{} and \limefull{} from
        full experiment from Section~\ref{sec:AIris} (repetition of
        Figure~\ref{fig:161_coeffs_airis} without the other methods)}
        \label{fig:161_coeffs_airis_fake}
    \end{subfigure}%
    \hfill
    \caption{Comparison of explanations from Section~\ref{sec:AIris} to
    explanations from Appendix~\ref{app:tabairis}}
    \label{fig:comparison_aIris_true}
\end{figure*}

\paragraph{Experiments and Results}

We compare \dbasimple{} and LIME to explain the true class boundary,
with settings as similar to Section~\ref{sec:AIris} as possible. Before
running the explanation methods, we standardize all features as
described in Appendix~\ref{app:standardizinghyperplanes}. For
\dbasimple{} we set $k = 1000, m = 500$ and $\rgrid =
\{0.1,0.2,\ldots,0.9,1,1.5,2,\ldots,9.5,10\}$. We instantiate LIME with
$m =500$ and $\sigma = 0.75 \sqrt{d}$ for $d = 5$. There is one notable
limitation: the true class assignments are deterministic, so we are
feeding LIME with class probabilities that are always either $0$ or $1$.
This is allowed for LIME \cite[Section~3.2]{LIME}, but it is different
from our approach in Section~\ref{sec:AIris} where we were feeding LIME
the non-deterministic class probabilities produced by a CNN.

Figure~\ref{fig:comparison_aIris_true} shows the explanations of DBA and
LIME for a single data point. It compares the results when running
\dbasimple{} and LIME directly on the latent parameters (simplified
experiment from current section) to running \dbafull{} and \limefull{}
on the corresponding image (full experiment from
Section~\ref{sec:AIris}). The results for DBA are very similar in both
cases: it recovers the true coefficients well. Performance appears to be
slightly better in Figure~\ref{fig:161_latentspace_airis_true} compared
to Figure~\ref{fig:161_coeffs_airis_fake}, as might be expected given
the simplified setup in which the latent parameters are directly
accessible. For LIME, we also see strong similarities between the two
figures. Surprisingly, its performance appears to be slightly worse in
the simplified setting of Figure~\ref{fig:161_latentspace_airis_true}
compared to the harder case of Figure~\ref{fig:161_coeffs_airis_fake}.
As we will see below, this is not representative of its general behavior
when explaining other cases.

\begin{table}[htb]
  \caption{Simplified Tabular AIris evaluation statistics, averaged over
  $50$ test points}
  \label{tab:global_measures_airis_tab}
  \centering
  \footnotesize
  \begin{tabular}{llllllll}
    \toprule
                    & DBA      & LIME           & Global    & Class & Decision Boundary       & Cosine         & Cosine         \\
                    & Fidelity & $R^2$-Fidelity & Fidelity  & Balance         & Distance     & Similary-      & Similarity+   \\
  \midrule
 \dbasimple{}         &  95\%  & -              &    -      & \textbf{50.1\%} & \textbf{0.7} &\textbf{0.906} & \textbf{0.998} \\
 LIME        &        - & 33.9\%         &    -      & 51.2\%          & 0.9          &  0.665        & 0.773 \\
    \bottomrule
  \end{tabular}
\end{table}

Aggregate results are reported in
Table~\ref{tab:global_measures_airis_tab}, which is the analogue of
Table~\ref{tab:global_measures_airis} from the full experiment,
evaluated on the same 50 test cases. For both methods we see that the
Cosine Similarities are similar to the results from
the full experiment, with small improvements between 0.05 and 0.09.
Fidelity is slightly down for DBA but still high. For LIME fidelity is
still low, but much better than in Section~\ref{sec:AIris}. It is not
clear that the distance to the decision boundary in
Table~\ref{tab:global_measures_airis_tab} can be directly compared to
the Latent Distance in Table~\ref{tab:global_measures_airis}, but they
are comparable nevertheless. Based on the similarities with the results
from Section~\ref{sec:AIris}, we conclude that the results in the
full AIris experiment are driven for a large part by the behavior of the
explanation methods and not, for instance, by peculiarities of the CNN
or VAE.

\subsubsection{UCI Heart Disease Data: Opposite Conclusions from DBA and
LIME}
\label{app:tabularUCI}

In this experiment we use the heart disease data from the UCI repository
\cite{UCI} to give a real-world example of a case where \dbasimple{} and
LIME lead to opposite conclusions on the importance of one of the
features. This shows that it can really make a difference whether we
explain the local decision boundary or the classifier probabilities
around $\x_0$. 

\begin{figure*}[htb]
    \centering
    \begin{subfigure}[t]{0.5\textwidth}
        \centering
        \includegraphics[width=0.88\textwidth]{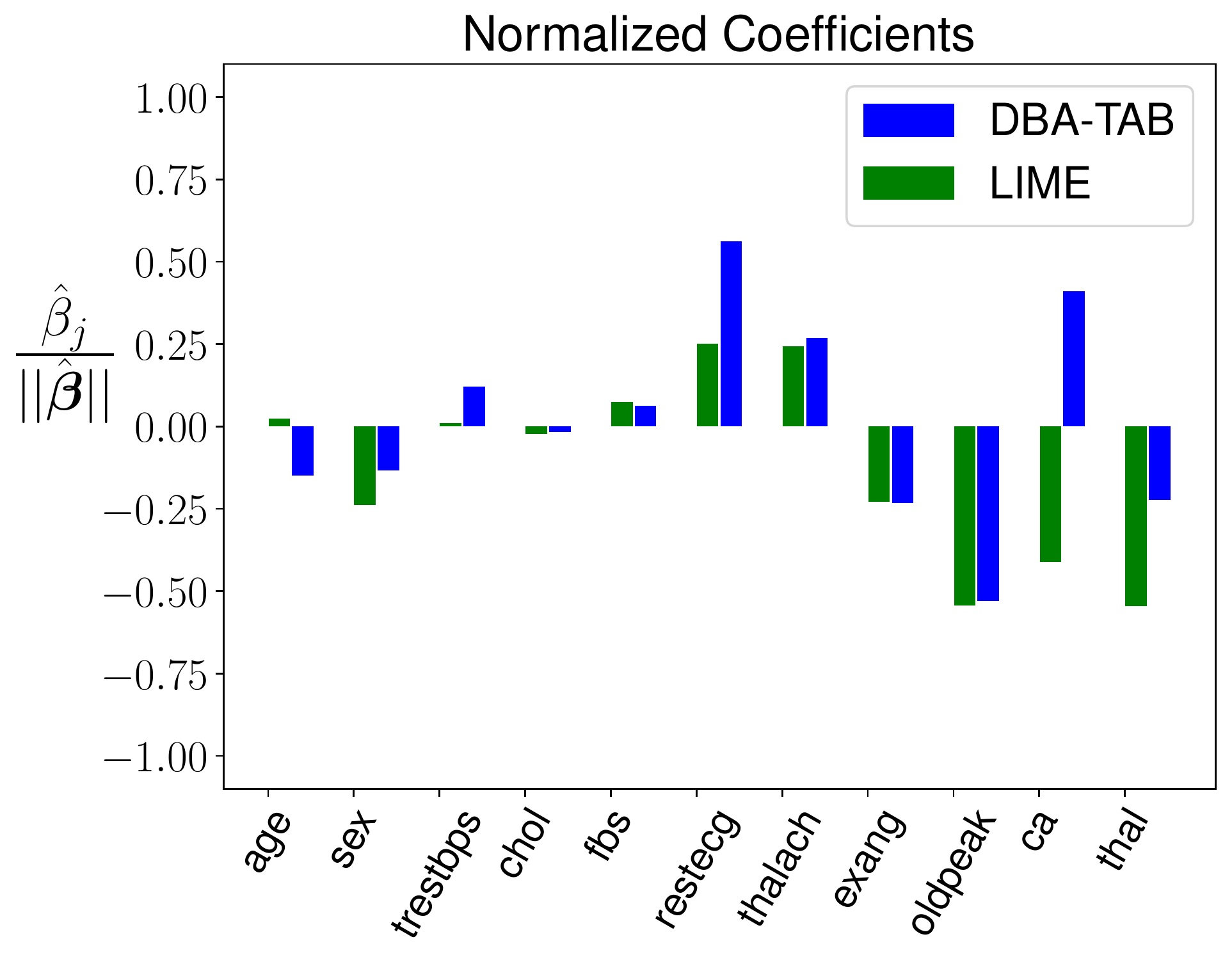}%
        \caption{Explanations}
        \label{fig:barplot_heart}
    \end{subfigure}%
    \begin{subfigure}[t]{0.5\textwidth}
        \centering
        \includegraphics[width=0.7\textwidth]{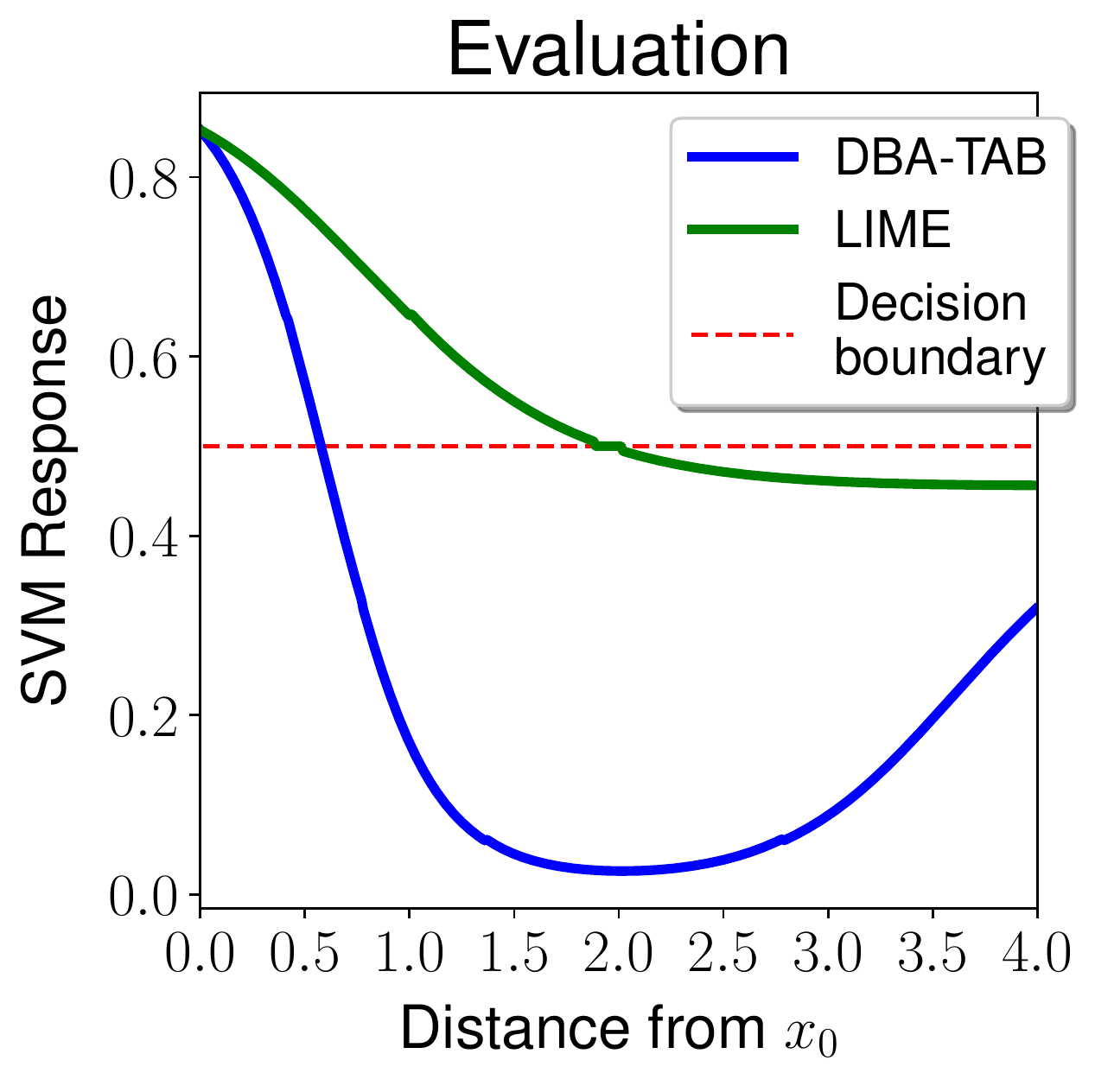}
        \caption{SVM probabilities as we move in direction
    $\estbeta/\|\estbeta\|$ from $\x_0$}
        \label{fig:path_heart}
    \end{subfigure}
    \caption{Comparison of \dbasimple{} and LIME on a single patient
    from the heart disease data}
    \label{fig:comparison_heart}
\end{figure*}

The heart disease data consist of 303 instances of patients that are
labeled according to whether they suffered a heart disease or not. There
are 13 features of mixed types, describing the health conditions of the
patients: age, sec, cp, trestbps, chol, fbs, restecg, thalach, exang,
slope, oldpeak, ca, thal. Since the linear surrogate models of both
\dbasimple{} and LIME become difficult to interpret on categorical
features with more than two possible values, we simplify the setting as
follows: we merge categories for features restecg and thal to make them
binary, and we omit features cp and slope for which there is no natural
way to merge categories. In case of restecg we merge categories 1 and 2,
which both code for an abnormal electrocardiographic measurement, and in
case of thal we merge categories 6 and 7, which both code for defects
related to a blood disorder called thalassemia. All resulting features
were standardized based on their means and standard deviations, which
were estimated on all available data. We randomly split the data into a
train set (242 instances) and a test set (61 instances), and train an
SVM with the same parameters as in the Moons experiment from
Appendix~\ref{app:moons}, to reach a test accuracy of 85.24\%. As in the
Moons data set, we obtain probabilistic classifications by Platt
scaling, which match the original SVM classifications in 99.6\% of the
cases in the whole data (train and test set together).

\paragraph{Experiments and Results}

Parameters for \dbasimple{} are the same as in
Section~\ref{sec:experiments} and Appendix~\ref{app:tabairis}. For LIME,
we set the kernel width equal to $\sigma = 0.75\sqrt{d}$ for dimension
$d=11$ and generate $m = 500$ samples.

Figure~\ref{fig:barplot_heart} shows the resulting explanations on a
single patient. We see that the explanations generally agree, except
that \dbasimple{} considers feature ca (the number of major vessels
(0-3)) to have a large positive influence on the probability of heart
disease, whereas LIME considers the same feature to have a large
negative influence. This of course is possible, because the two
explanation methods have different goals: explaining the local decision
boundary versus explaining the classifier probabilities around $\x_0$.
Figure~\ref{fig:path_heart} shows that, indeed, the LIME explanation is
much worse than \dbasimple{} at indicating the direction to the closest
decision boundary point.
So in this case the two goals are incompatible.

\begin{table}[h]
  \caption{Heart disease evaluation statistics, averaged over the whole
  data set (both train and test set)}
  \label{tab:global_measures_heart}
  \centering
  \footnotesize
  \begin{tabular}{llllll}
    \toprule
                    & DBA      & LIME            & Class           & Decision Boundary\\
                    & Fidelity & $R^2$-Fidelity  & Balance         & Distance  \\
  \midrule
 \dbasimple{}         &  94.3\%  & -           & \textbf{49.2\%} & \textbf{1.18}  \\
 LIME        &        - & 46.4\%         &     45.8\%          & 1.30          \\
 
    \bottomrule
  \end{tabular}
\end{table}

A natural follow-up question is whether it is typical that the
difference between \dbasimple{} and LIME is this large. As a small
comfort, it turns out the distances to decision boundary are on average
much more similar, as can be seen in the aggregate statistics from
Table~\ref{tab:global_measures_heart}. 

\subsubsection{Additional Experiments on UCI Tabular Data Sets}

Here we compare \dbasimple{} to both LIME and MAPLE on four additional
standard UCI data sets: Abalone, Breast Cancer:Diagnosis, Arrhythmia and
Accent Recognition MFCC. For \dbasimple{} we use the same settings as in
Section~\ref{sec:experiments} and Appendix~\ref{app:tabairis}; we apply
LIME with $\sigma = 0.75\sqrt{d}$ and $m = 500$; and for MAPLE we use a
random forest with 200 trees and half of the features considered at each
split. This random forest is used for supervised neighborhood selection
\cite{MAPLE,MAPLE-code}. For all explanation algorithms we use an
unpenalized explainer (linear regression for LIME and MAPLE, logistic
regression for \dbasimple{}).

\paragraph{Data and Models}

We randomly split all datasets in train (90\%) and test sets (10\%).
For the Abalone data ($d = 9$) we binarize the target (number of rings)
by predicting whether or not it is above the threshold $t = 10$. We
train a neural network (3 hidden layers, 10 neurons each) to reach a
test accuracy of 83.4\%. For the Breast Cancer data ($d =32$) we treat
sex as a binary numerical variable for all algorithms and we train a
Random Forest with 100 trees to classify tumors as either malignant or
benign with 95\% classification accuracy on the test set. In Arrhythmia
($d = 279)$ we exclude column 13 (QRST), which has mostly missing
values, and we further binarize the target in two (coincidentally
balanced) classes: normal versus arrhythmia. An Adaboost classifier with
70 rounds of boosted decision stumps yields 84.6\% test accuracy. Last,
for the Accent Recognition data ($d = 32$) a Neural Network with a
single hidden layer of 100 neurons learns to distinguish US speakers
from Europeans with 81.9\% test accuracy. Before training, all features
in all data sets were standardized based on their mean and standard
deviation, which were estimated on all available data.

\paragraph{Experiments and Results}

Table~\ref{tab:real_datasets_aggregate} lists evaluation statistics for
the three explanation algorithms, averaged over the test set of each
data set. We see that all methods score lowest on average fidelity for
the Abalone data. Fidelity is especially high for \dbasimple{} on
Arrhythmia and Accent, and for MAPLE on Arrhythmia. In terms of the
distance to the decision boundary when moving along the proposed
explanation direction, we see a clear ordering: \dbasimple{} is best in
all cases, followed by LIME and then MAPLE, which is significantly
worse. In fact, for a significant percentage of the explained points in all data sets (except for Accent),
moving from $\x_0$ in the direction of the explanation proposed by MAPLE
did not cross the decision boundary at all, even in cases where its
fidelity to the class probabilities was high. We therefore include an
extra column that counts such failures to cross the decision boundary,
and we average Decision Boundary Distance only for the cases in which
none of the algorithms failed. For Arrhythmia and Breast Cancer, LIME
also sometimes fails, but with smaller rates than MAPLE, while
\dbasimple{} never fails. We conclude that \dbasimple{} is much better
at approximating the nearest decision boundary direction than LIME and
MAPLE, which is of course not surprising, because LIME and MAPLE have
been designed for a different purpose, namely to approximate the class
probabilities around $\x_0$.

\newcolumntype{Y}{>{\raggedleft\arraybackslash}X}%

\begin{table*}[htb]
  \centering
  \caption{Evaluation statistics for UCI tabular data sets, averaged
  over their test sets}
  \label{tab:real_datasets_aggregate}
  \begin{tabularx}{\linewidth}{l*{6}{Y}}
    \toprule
                    & DBA Fidelity  & $R^2$ Fidelity & Class Balance &
                    Decision Boundary Distance  & Failure to Cross
                    Decision Boundary\\
  \toprule
  \toprule
  \multicolumn{6}{l}{\textbf{Abalone}} \\                    
  \midrule
 DBA-TAB        &  80.2\%  & -           & \textbf{49.7\%} & \textbf{0.26} & \textbf{0.0 \%}\\
 LIME        &        - & 66.7\%         &     40.1\%          & 0.30         &  \textbf{0.0 \%}\\
 MAPLE       &        - & 46.2\%         &     -          & 0.38&     10.1 \%      \\
    \bottomrule
    \toprule
    \multicolumn{6}{l}{\textbf{Breast Cancer}} \\
  \midrule
 DBA-TAB        &  81.3\%  & -           & 56.2\% & \textbf{2.03} &  \textbf{0.0 \%}\\
 LIME        &        - & 79.1\%         &    \textbf{47.9}\%          & 2.57  & 2.2 \% \      \\
 MAPLE       &        - & 84.4\%         &     -          & 4.68 & 29.3\%         \\
    \bottomrule
    \toprule
    \multicolumn{6}{l}{\textbf{Arrhythmia}} \\
  \midrule
 DBA-TAB        &  82.0\%  & -           & \textbf{49.2\%} & \textbf{1.03} & \textbf{0.0 \%} \\
 LIME        &        - & 61.1\%         &     29.7\%          & 2.92        & 9.9 \% \\
 MAPLE       &        - & 94.7\%         &      -          & 8.92 &   5.0 \%      \\
    \bottomrule
    \toprule
    \multicolumn{6}{l}{\textbf{Accent MFCC}} \\
  \midrule
 DBA-TAB        &  95.5\%  & -           & \textbf{50.5\%} & \textbf{0.89} & 0.0 \%  \\
 LIME        &        - & 84.4\%         &     45.8\%          & 0.99        & 0.0 \%  \\
 MAPLE       &        - & 68.5\%         &    -         & 1.15         & 0.0 \% \\
    \bottomrule
  \end{tabularx}   
\end{table*}

\subsection{Additional Details for \dbafull{}: Modification to VAE
Training}
\label{app:VAEtrainingstability}

As mentioned in Section~\ref{sec:dbafull}, we adjust the training
procedure of the VAE to favor preservation of class probabilities. Let
$c(\x)$ be the probability that a classifier $c$ assigns to class $+1$.
(This can be replaced by binary classifications $f(\x)$ if no
probabilities are available.) Then what we want is for $c(\x)$ and
$c(\x')$ to be as close as possible when $\x'$ is the result of mapping
$\x$ to the latent space and back using the VAE. During the training of
the VAE we therefore monitor the stability of the probabilities on a
hold-out set:
\begin{equation}\label{eqn:probabilitystability}
  \frac{1}{n} \sum_{i=1}^n |c(\x_i) - c(\x'_i)|.
\end{equation}
We initially tried to add this stability as an extra term to the
objective function that is being minimized to train the VAE, but we
found that this was harmful to the linearity of the resulting latent
space. We therefore settled on an alternative solution, which was to
keep the standard training procedure, but to calculate the
\emph{probability stability} \eqref{eqn:probabilitystability} once per
epoch and finally output the VAE parameters that minimize
\eqref{eqn:probabilitystability} during training. This significantly
improved label stability both for AIris and for CelebA, without harming
the linearity of the latent space.

\section{Other Methods}
\label{app:other_methods}
In this section, we provide the details of how we implemented the
\limefull{} and CEM-MAF methods.

\subsection{\limefull{}}
\label{app:limefull}

We provide a variant of LIME \cite{LIME} and ALIME \cite{ALIME} that
works in the latent space of a VAE and can explain based on
user-supplied annotations, in the same manner as \dbafull{}. This
procedure, which we call \limefull{}, can be viewed as a modification of
\dbafull{} in which we replace the \dbasimple{} part by LIME. We keep
the VAE and annotators the same as in \dbafull{}, so we can use the same
latent space and attribute space, as shown in
Figure~\ref{fig:attdba_scheme}.

The sampling procedure described in the LIME paper \cite{LIME} is not
applicable when sampling in the latent space, because it requires
mappings back and forth between the vectors $\z$ and an interpretable
binary representation, which is not available in the latent space. We
therefore use the approach used in the LIME software
\cite{LIMESoftware2020} to sample from continuous features, as already
described in Appendix~\ref{app:moons}, except that now we apply it to
the latent representations $\z$ instead of the inputs $\x$. The sample
points are then augmented with their corresponding predicted
probabilities, which are obtained through the mapping $\z \mapsto \x
\mapsto c(\x)$. \limefull{} further measures distance in the latent
space: the weight of a sample point $\z$ is $\pi_{z_0}(\z) =
\exp{(-\|\z-\z_0\|^2/\sigma^2)}$ where $\z_0$ is the latent space
representation of $\x_0$. Following the LIME software
\cite{LIMESoftware2020}, we always set $ \sigma = 0.75\sqrt{l}$ where
$l$ is the dimensionality of $\latentspace$. Finally $\limefull{}$ fits
a linear surrogate model $g$ using weighted least squares in
$\attributespace$ by mapping the generated samples $\z$ to corresponding
attribute vectors $\a$ using the annotators. The attributes of $a_j$ in
$\a$ are standardized based on their means and standard deviations in
the \limefull{} sample.

A notable difference between our approach and the LIME paper \cite{LIME}
is that we do not impose sparsity of the linear surrogate model $g$ in a
pre-selection step with the LASSO, which is not needed because the
number of user-specified attributes is small enough to be interpretable
in all our experiments. The LIME software \cite{LIMESoftware2020}
further offers the possibility to add $L_2$-penalization, which we omit
for the same reason.

\subsection{CEM-MAF}
\label{app:cem_maf}

The Contrastive Explanation Method with Monotonic Attribute Functions
(CEM-MAF) \cite{CEM-MAF} is an extension of CEM \cite{CEM}, which
explains the prediction for $\x_0$ by generating a contrastive example
$\xcontrast$ that is similar to $\x_0$ but differs in an informative
way. In the CEM framework there are two types of contrastive examples.
The first type is \emph{pertinent positive}, in which case
$\xcontrast$ corresponds to removing as many features from $\x_0$ as
possible while maintaining the same class: $f(\xcontrast) = f(\x_0)$.
The second type is \emph{pertinent negative}, which means that
$\xcontrast$ corresponds to adding as few features to $\x_0$ as possible
in order to change the class $f(\xcontrast) \neq f(\x_0)$. Although CEM
and CEM-MAF do not produce pertinent negatives that lie on the decision
boundary, the idea is similar in spirit to our DBA approach of
identifying the fastest direction from $\x_0$ to get to a point on the
decision boundary. Other similarities are that CEM-MAF can learn
user-specified attributes from user annotations and can measure distance
in the latent space of a VAE. We therefore compare to the CEM-MAF
pertinent negative method.

CEM-MAF requires a VAE (or GAN) to map back and forth between inputs
$\x$ and latent representations $\z$. In our experiments we use the same
VAE for both \dbafull{} and CEM-MAF for a fair comparison. Let $c(\x)$
be the probability that $\x$ is in class $+1$ according to a
probabilistic binary classifier $c$. Then CEM-MAF for pertinent
negatives minimizes the following objective:
\begin{equation}\label{eqn:CEM_MAF_loss}
    \min_{\zcontrast \in \inputspace}\quad 
    F_{C,\kappa}(\xcontrast)
    + \eta \|\xcontrast - \x_0\|_2^2 + \nu \|\zcontrast - \z_0\|_2^2
    + A_{\gamma,\mu}(\zcontrast),
\end{equation}
where $\xcontrast$ is the reconstruction in input space that corresponds
to the latent representation $\zcontrast$. Here the term
$F_{C,\kappa}(\xcontrast)$ encourages $\xcontrast$ to be classified as
the opposite class of $\x_0$. If $c(\x_0) \geq 0.5$, i.e.\ $\x_0$ is
classified as class $+1$, it is defined as
\[
  F_{C, \kappa}(\xcontrast)
    = C \cdot \max \Big \{c(\xcontrast) - (1 - c(\xcontrast)),
    -\kappa\Big \}
\]
If $\x_0$ is classified as $-1$, then both occurrences of
$c(\xcontrast)$ should be replaced by $1-c(\xcontrast)$. The second and
third term in \eqref{eqn:CEM_MAF_loss} respectively minimize the
distance between $\x_0$ and $\xcontrast$ in the input space and in the
latent space. Finally, the last term enforces addition (and not removal)
of interpretable attributes:
\begin{equation*}
 A_{\gamma ,\mu}(\zcontrast)
  = \gamma \sum_{i=1}^p{\max{\{h_i(\x_0) - h_i(\xcontrast), 0\}}}
    + \mu \sum_{i=1}^p |h_i(\xcontrast)|.
\end{equation*}
There is a function $h_i : \inputspace \to \reals$ for each interpretable attribute
$i=1,\ldots,p$. A higher value $h_i(\x)$ indicates stronger presence of
attribute $i$ in input $\x$, and in typical usage they range from $0$ to
$1$. Each function $h_i$ is learned from user-supplied annotations by a
separate neural network. Thus these functions are similar to the
annotators in \dbafull{}, except that they are non-linear and operate in
the input space instead of the latent space.

The objective \eqref{eqn:CEM_MAF_loss} is non-convex. Optimization with
standard stochastic optimization procedures therefore does not always
give the same solution. It depends on hyperparameters $C$, $\kappa$,
$\eta$, $\nu$, $\gamma$ and $\mu$, which the user needs to tune to
optimize convergence. An automatic tuning procedure is available for
$C$, which increases computational cost substantially. The algorithm
also bears the unavoidable computational cost of the training and tuning
of $p$ (convolutional) neural networks with many parameters, to serve as
annotators. By comparison, the annotators in \dbafull{} take much less
computation and fewer user-annotations, because they only require
fitting linear models with logistic regression.

Tuning the hyperparameters for CEM-MAF is non-trivial and requires
elaborate computationally expensive experimentation. We select
hyperparameters for each experiment individually to guarantee that the
optimizer always converges to a small objective value. To make
experiments for many explanations and algorithms computationally
feasible, we do not tune $C$ automatically but instead find for each
experiment the smallest $C$ that yields a small objective value for all
inputs that we explain. We select values for the other hyperparameters
according to the authors' suggestions \cite{CEM-MAF} and by observing
convergence.

\section{AIris Experiment}
\label{app:airis}

In this section, we provide additional information about the AIris
experiment reported in Section~\ref{sec:AIris}. We first discuss how the
images were generated. Then, in
Section~\ref{app:standardizinghyperplanes}, we show how standardizing
the features changes the coefficients of the hyperplanes from
\eqref{eqn:hyperplanes}. In Section~\ref{sec:airisglobalmappings} we
provide details on how we trained the VAE, CNN and annotators.
Section~\ref{sec:airisothermethods} contains details for \limefull{},
CEM-MAF and \globalsurrogate{}. And finally,
Section~\ref{sec:airisadditional} shows additional results on the best
and worst performance of \dbafull{} as well as a PCA projection of its
sample $S$ onto two dimensions.

\subsection{Image Generation: Details and Examples}

The image generation program for the AIris data generates images of
flowers using third-order connected Bezier curves. It takes as input the
parameters from Table~\ref{tab:airisparams}, and creates flowers with an
equal number of petals and sepals using standard open source
visualization software \cite{matplotlib}. In order for the flowers to be
realistic we had to bound the ranges of their shape parameters in the
intervals shown in Table~\ref{tab:airisparams}. Color of the petals is
manipulated by a continuous mixing parameter C that interpolates between
red and magenta.

\begin{table}[htb]
  \centering
  \caption{AIris parameters}
  \label{tab:airisparams}
  \begin{tabular}{rcccc}
    \toprule
        & & range       & mean & sd   \\
    \midrule
    Petal length & \PL & $[0.3,0.7]$ & 0.5  & 0.115\\
    Petal width  &\PW & $[0.1,0.7]$ & 0.4  & 0.173\\
    Sepal length & \SL & $[0.3,0.7]$ & 0.5  & 0.115\\
    Sepal width  &\SW & $[0.1,0.7]$ & 0.4  & 0.173\\
    Color        &\C  & $[0.1,0.8]$ & 0.45 & 0.202\\
    \bottomrule
  \end{tabular}
\end{table}

\begin{figure*}[htb]
    \centering
    \hfill
    \begin{subfigure}[t]{0.4\textwidth}
        \centering
        \includegraphics[width=1\textwidth]{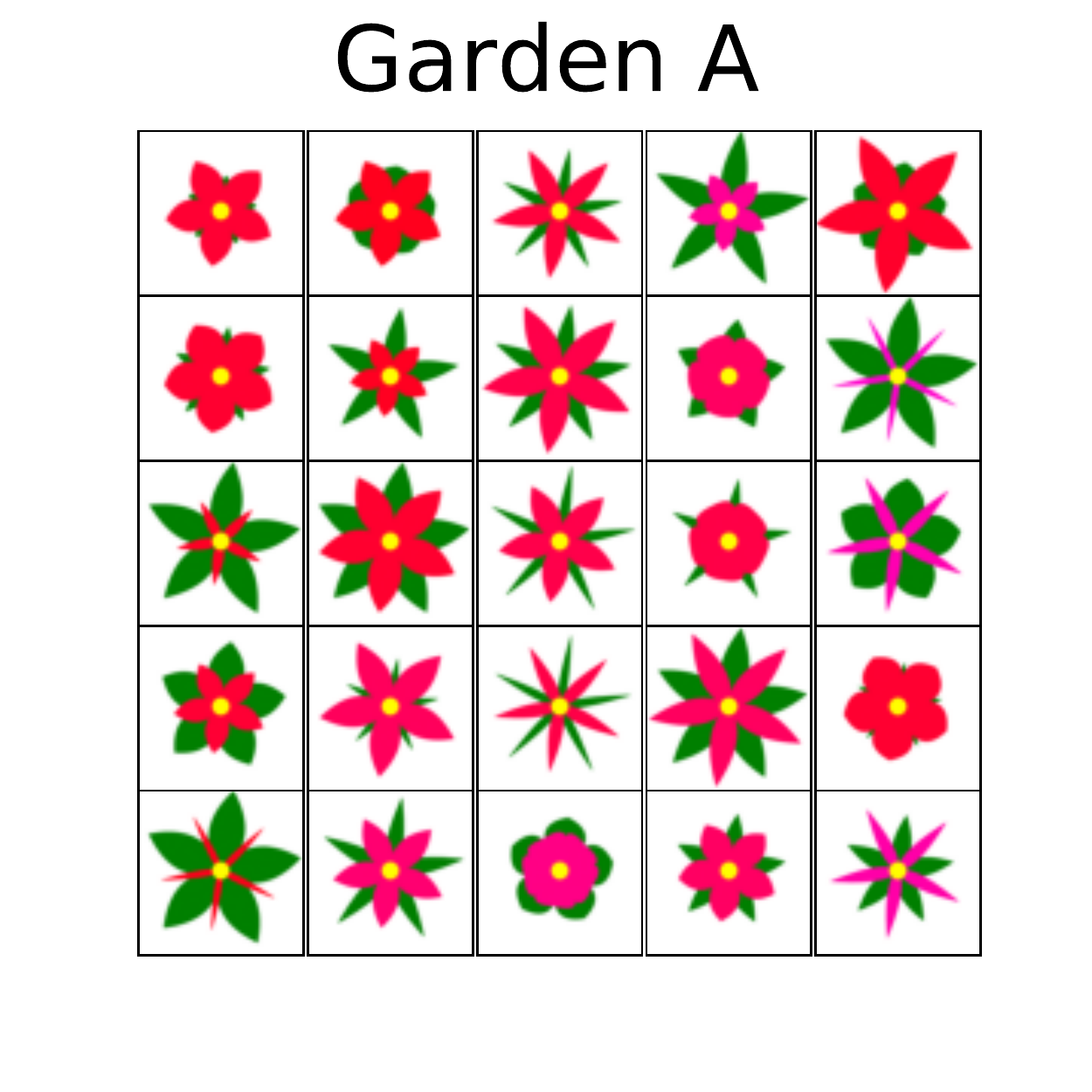}%
        \caption{Examples from class A}
        \label{fig:gardenA}
    \end{subfigure}%
    \hfill
    \begin{subfigure}[t]{0.4\textwidth}
        \centering
        \includegraphics[width=1\textwidth]{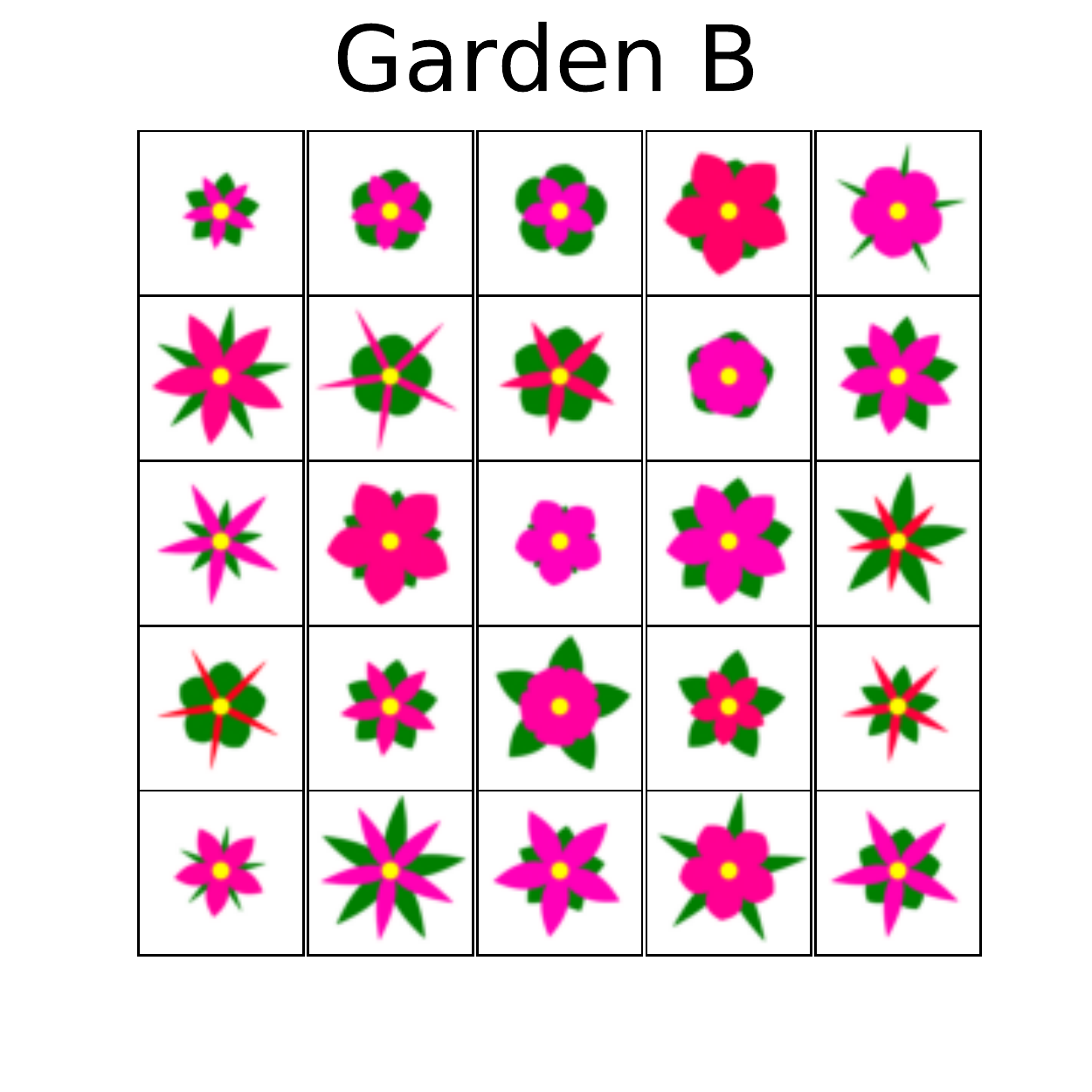}
        \caption{Examples from class B}
        \label{fig:gardenB}
    \end{subfigure}%
    \hfill
    \caption{Examples of AIris data}
    \label{fig:gardens}
\end{figure*}

Flowers are generated by uniformly sampling parameters over their range
of allowed values, and splitting them into classes A and B according to
the non-linear rule \eqref{eqn:hyperplanes}. We sample 4000 images for
training as well as 2000 test images. The proportion of class A in the
training set is 47.43\%, so the two classes are approximately balanced.
Figure~\ref{fig:gardens} shows examples for the two resulting
``gardens''. A first impression is that color (C) separates the classes,
which is true since C is an non-zero coefficient of the first
hyperplane. However more features are responsible for separating the
gardens and it is hard for a human eye to capture the ground truth.

\subsection{Standardizing Features and Coefficients of the True Hyperplanes}
\label{app:standardizinghyperplanes}

As mentioned in a footnote~\cref{foot:hyperplane} in
Section~\ref{sec:AIris}, the coefficients of the hyperplanes from
\eqref{eqn:hyperplanes} need to be standardized before comparing them to
the explanations produced by the explanation methods. This works as
follows: we first consider standardized features $\sPL, \sPW, \sSL,
\sSW, \sC$, which are obtained from the parameters by subtracting the
mean and dividing by the standard deviation of the parameter. The ranges
for the parameters are specified in Table~\ref{tab:airisparams}, which
also shows their means and standard deviations, where we use that the
standard deviation of a uniform distribution on an interval $[a,b]$ is
$(b-a)/\sqrt{12}$. The true decision boundary from
\eqref{eqn:hyperplanes} then becomes the rule to assign to class~A if
\begin{equation*}
  0.038\sPL + 0.057\sPW + 0.067\sC < .0545   %
  \quad \text{ and } \quad
  0.038\sPL + 0.057\sPW + 0.038\sSL > -.062, %
\end{equation*}
In particular, the coefficients of the hyperplanes are obtained from the
unstandardized coefficients by multiplying by the standard deviations of
the corresponding features.

\subsection{Global Mappings: VAE, CNN and Annotators}
\label{sec:airisglobalmappings}

Here we provide details on how we trained the VAE, CNN and annotators
for the AIris experiment. These are the global mappings between spaces
in Figure~\ref{fig:attdba_scheme}.

\subsubsection{VAE}

We trained a convolutional variational autoencoder (CVAE) that is a
slight modification of the CVAE of \citet{DFCVAE}. The objective
function being optimized has two terms: reconstruction loss (implemented
with binary cross-entropy) and Kullback-Leibler divergence, weighted by
factors 0.5 and 1.2 respectively. The CVAE is implemented in TensorFlow
2.2. The latent space has 10 dimensions.

\paragraph{Architecture}

The architecture we used is a slight modification of the CVAE of
\citet{DFCVAE}. Our inputs and outputs are $128 \times 128$ RGB images.
The encoder part of the CVAE consists of 5 blocks with each block
consisting of a convolutional layer followed by a Batch Normalization
layer and a LeakyReLU layer ($\alpha = 0.3$). The convolutional layers
of the 5 blocks all have kernel size $4 \times 4$, a stride parameter
of 1 and filter sizes 32, 64, 128, 256 and 512 with that order. The 5
blocks are then followed by a global average pooling operation and a
dense layer with output dimension 20 that is interpreted as producing 10
means and 10 corresponding log-variances.
The decoder starts with
a dense layer of 4096 neurons and then mirrors the architecture of the
encoder with upsampling and nearest neighbor interpolation. Filter sizes
for the 5 convolutional layers are 344, 64, 32, 16 and 3, with kernel
size $3 \times 3$. The first 4 convolutional layers are also followed by
Batch Normalization and LeakyReLU operations. The last convolutional
layer has a sigmoid activation function, since we optimize for binary
cross-entropy. We also experimented with $MSE$ reconstruction loss but
achieve better attribute vector quality with the former so we report
results with binary cross-entropy reconstruction loss.

\paragraph{Training Details}

We use the Adam optimizer with learning rate $0.001$, $\beta_1=0.9$,
$\beta_2 = 0.999$ and zero decay. We train for 200 epochs with a batch
size of 500 and store the weights that minimize the probability
stability from \eqref{eqn:probabilitystability}, as discussed in
Appendix~\ref{app:VAEtrainingstability}.

\paragraph{Qualitative Evaluation}

\begin{figure*}[htb]
    \centering
    \begin{subfigure}[t]{0.4\textwidth}
        \centering
        \includegraphics[scale=0.4]{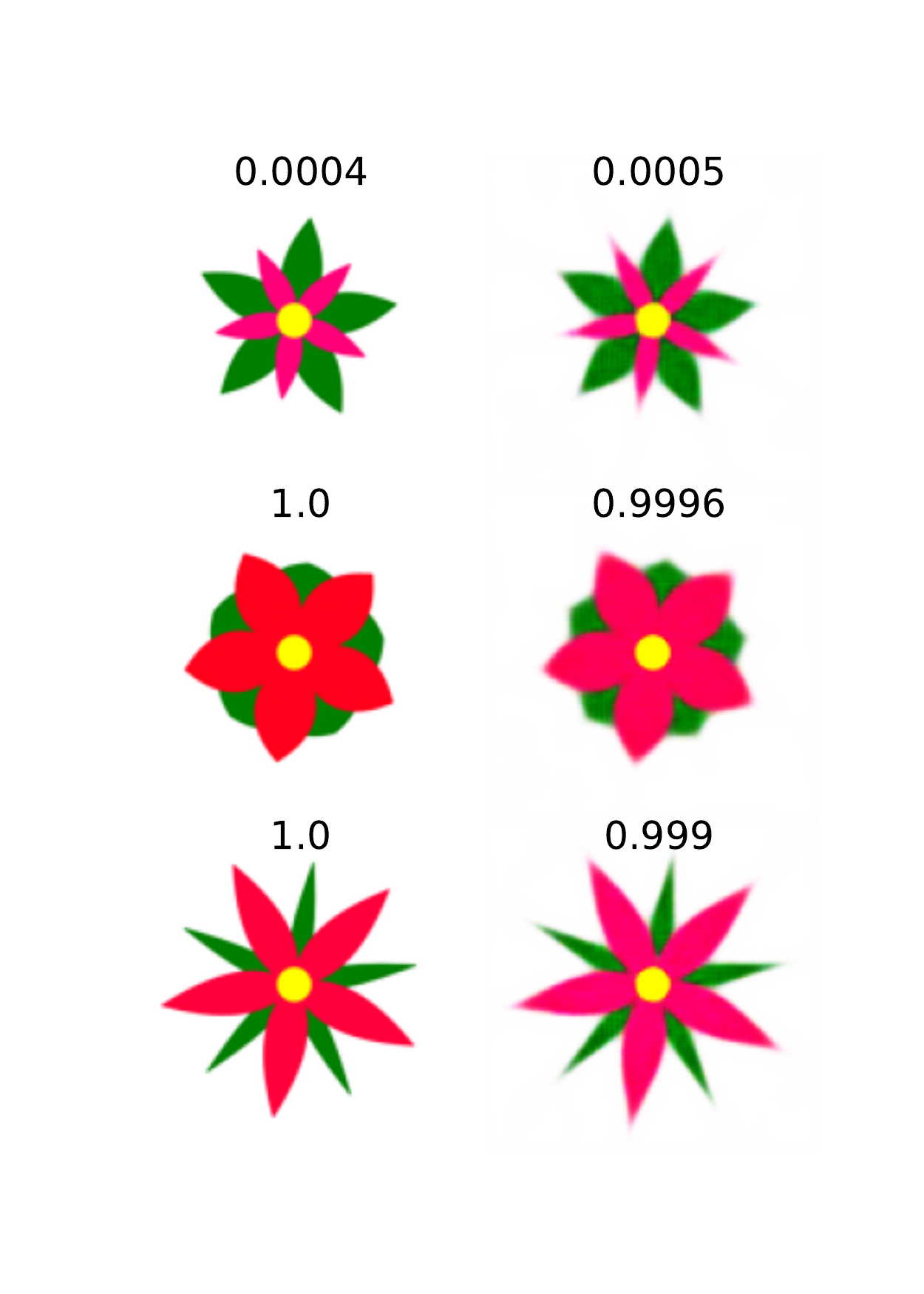}
       \caption{Original vs reconstruction}
        \label{fig:airisreconstructions}
    \end{subfigure}%
    ~ 
    \begin{subfigure}[t]{0.4\textwidth}
        \centering
        \includegraphics[scale=0.4]{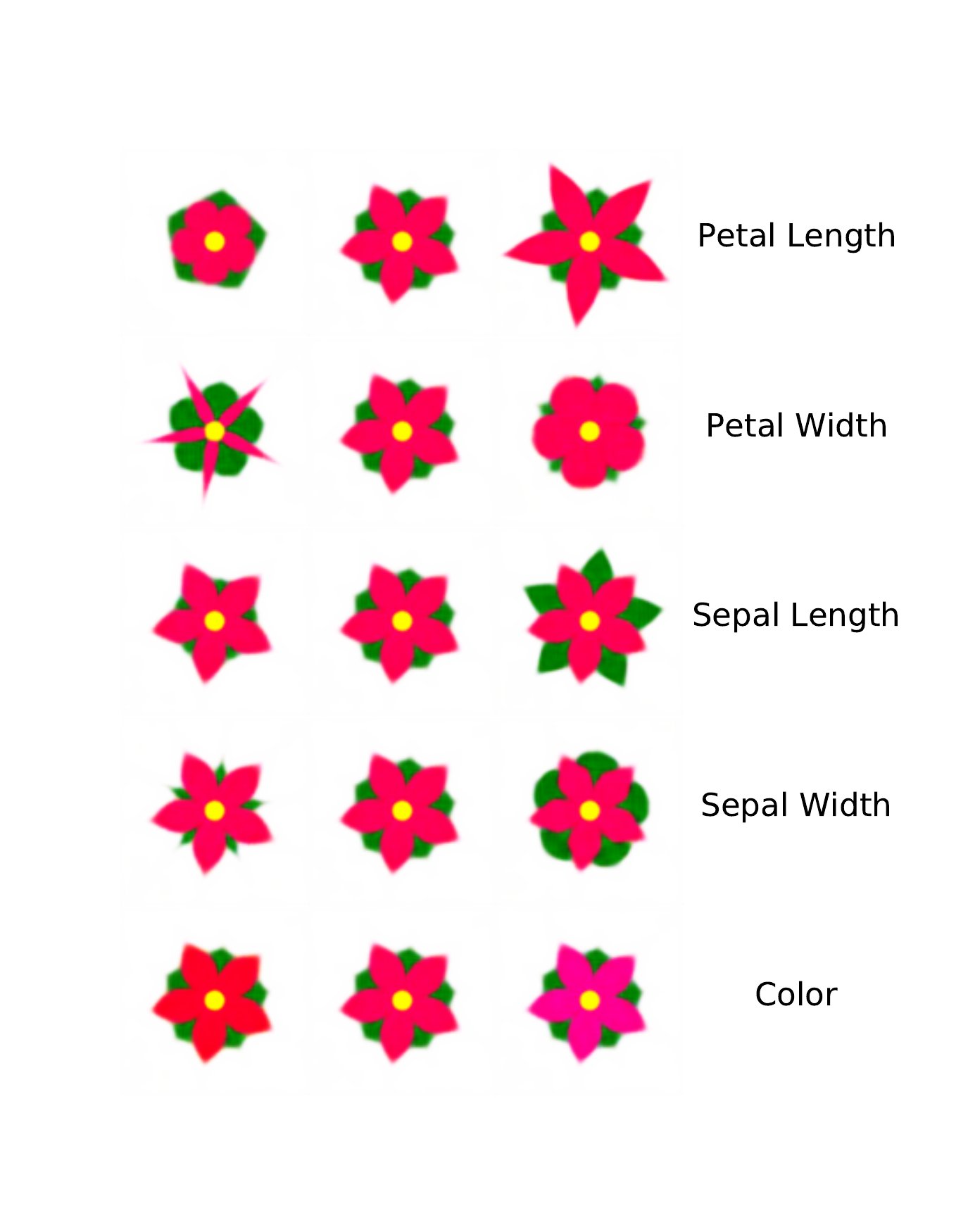}
        \caption{Quality of attribute vectors}
        \label{fig:airisatt_vecs}
    \end{subfigure}
    \caption{Qualitative evaluation of global mappings. On the left the
    original images and their CNN probabilities are compared with the
    reconstructions produced by the CVAE. On the right the original image
    (middle column) is varied along the direction of each learned
    attribute.}
    \label{fig:flowers}
\end{figure*}

Figure~\ref{fig:airisreconstructions} shows examples of the reconstruction
quality when mapping $\x \mapsto \z \mapsto \x'$ as well as the
corresponding CNN probabilities $c(\x)$ and $c(\x')$.

\subsubsection{CNN}

Our CNN architecture is almost the same as the encoder architecture of
the VAE, since we noticed good feature extraction behavior under this
design. What differs is that the last dense layer of the network
consists of 200 neurons. The output layer has a sigmoid activation
function and outputs the probability $c(\x)$ of $\x$ being of class A.
We optimize using stochastic gradient descent (SGD) with a learning rate of
$10^{-4}$. The CNN was trained for 200 epochs with a batch size of 128.
We use early stopping and store the weights in the epoch that validation
binary cross entropy loss is minimized (here epoch 156). This gives high
accuracy: 99.33\% and 98.75\% in the train and test set, respectively. 

\subsubsection{Annotators}

\dbafull{} annotators are $L_2$-penalized logistic regression models which learn
the mappings $\latentspace \to \attributespace$. 

Because of the danger of entanglement discussed in
Section~\ref{sec:discussion},
it is important to verify that the annotators indeed express the
concepts that the user had in mind. This can be seen in
Figure~\ref{fig:airisatt_vecs}: the middle column always shows the same
example image and the other two columns show the effect of changing one
of the attributes by a unit step in the latent space. Concretely, we map
the input image $\x$ to a latent representation $\z$. Then add (or
subtract) $\esttheta_j/\|\esttheta_j\|$
to $\z$
and map back to a corresponding input $\x'$, where $\esttheta_j$ are the
coefficients of the annotator for attribute $j$. The left column
corresponds to decreasing an attribute; the right column to increasing
it. 

\subsection{Details for \limefull{}, CEM-MAF and \globalsurrogate{}}
\label{sec:airisothermethods}

\paragraph{\limefull{}}

For a general description of \limefull{} see
Appendix~\ref{app:limefull}. We run \limefull{} with $m=500$ and kernel
width $\sigma = 0.75 \cdot \sqrt{10}$, because the dimension of the
latent space is $10$.

\paragraph{CEM-MAF}

For a general description of CEM-MAF, see Appendix \ref{app:cem_maf}. We
optimize its objective \eqref{eqn:CEM_MAF_loss} with SGD using a
constant learning rate of $0.0001$, which seems to yield stable
convergence. To save computation, we run 500 SGD iterations for all
explanations since CEM-MAF usually converged after about 400 iterations.
Before running the experiment for all methods we have tuned CEM-MAF
parameters empirically and according to the authors' suggestions
\cite{CEM-MAF}. We report results with the best set of parameters found,
which is $\{C , \kappa, \eta, \nu, \gamma , \mu \} = \{ 2500,
5,0.1,1,100,100\}$.

CEM-MAF returns $\xcontrast$ which is a contrastive example from the
opposite class of $\x_0$. Thus the corresponding latent direction
$\latentdirection = \zcontrast - \z_0$ points towards the decision
boundary in the latent space. It is also possible to obtain a
coefficient vector $\estbeta$ from CEM-MAF, which represents the
importance of various attributes in distinguishing $\xcontrast$ from
$\x_0$. Following its authors \cite{CEM-MAF}, we obtain this vector as
\[
  \estbeta = 
    \begin{pmatrix}
      h_1(\xcontrast) - h_1(\x_0)\\
        \vdots \\
      h_p(\xcontrast) - h_p(\x_0)
    \end{pmatrix}.
\]
We use this vector as the final CEM-MAF explanation in
Figure~\ref{fig:161_coeffs_airis} as well as to compute Cosine
Similarity+ and Cosine Similarity- measures in
Table~\ref{tab:global_measures_airis}.

\paragraph{\globalsurrogate{}}

Finally, \globalsurrogate{} is trained on the training data with access
to the original latent parameters PL, etc. The latent parameters 
are standardized based on their empirical means and standard deviations.
We then use unpenalized logistic regression to fit a global surrogate
model to predict the class labels of the CNN.

\subsection{Additional Results: Best Case, Worst Case and PCA}
\label{sec:airisadditional}

\begin{figure*}[htb]
    \centering
    \begin{subfigure}[t]{0.55\textwidth}
        \centering
        \includegraphics[width=0.55\textwidth]{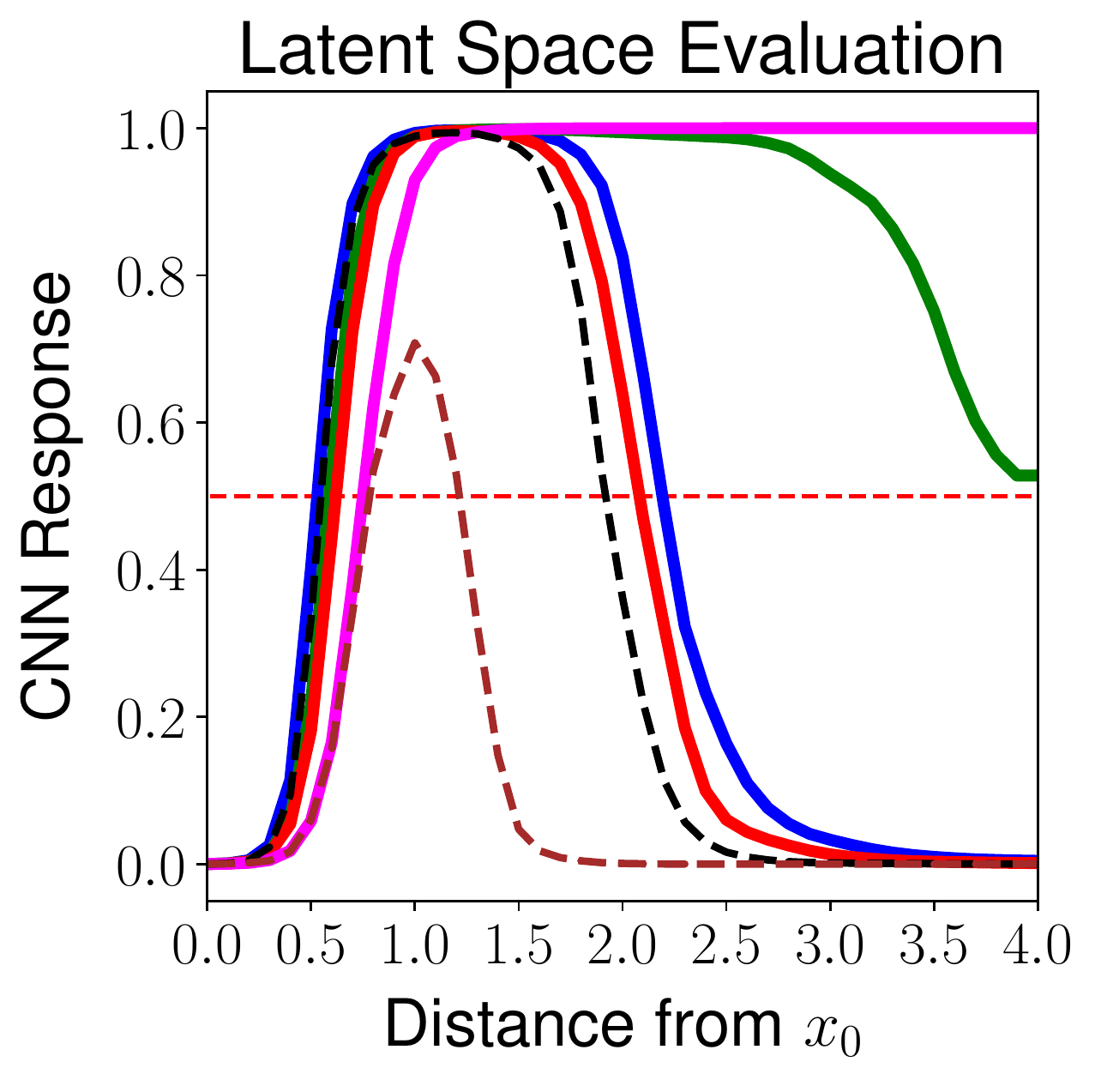}%
        \caption{CNN probabilities moving in direction
        $\latentdirection/\|\latentdirection\|$ from $\x_0$}
        \label{fig:141_latentspace_airis}
    \end{subfigure}%
    \begin{subfigure}[t]{0.45\textwidth}
        \centering
        \includegraphics[width=0.94\textwidth]{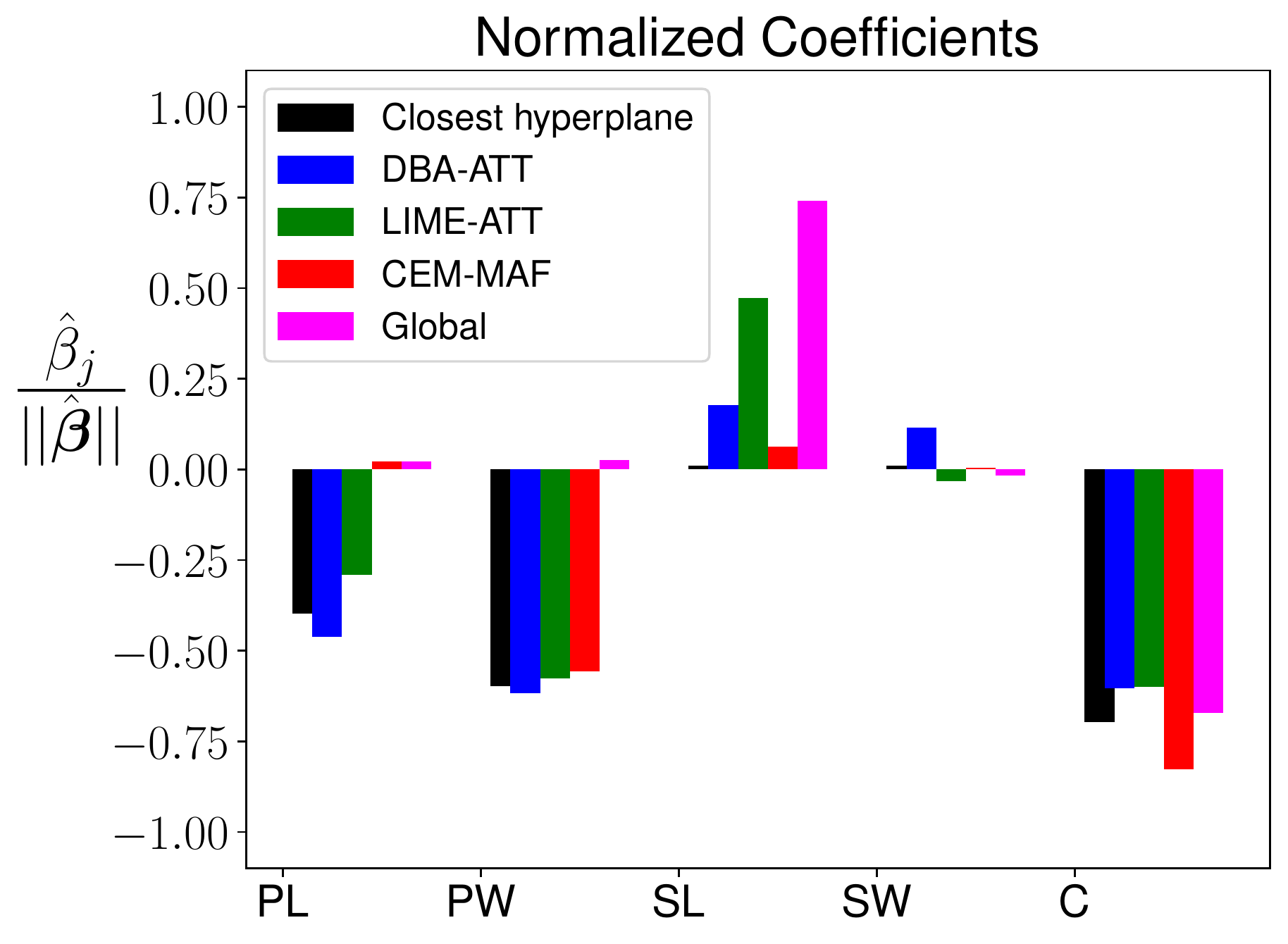}
        \caption{Explanation coefficients $\estbeta$}
        \label{fig:141_coeffs_airis}
    \end{subfigure}
    \caption{Worst-case example}
    \label{fig:141_aris}
\end{figure*}

\begin{figure*}[htb]
    \centering
    \begin{subfigure}[t]{0.55\textwidth}
        \centering
        \includegraphics[width=0.55\textwidth]{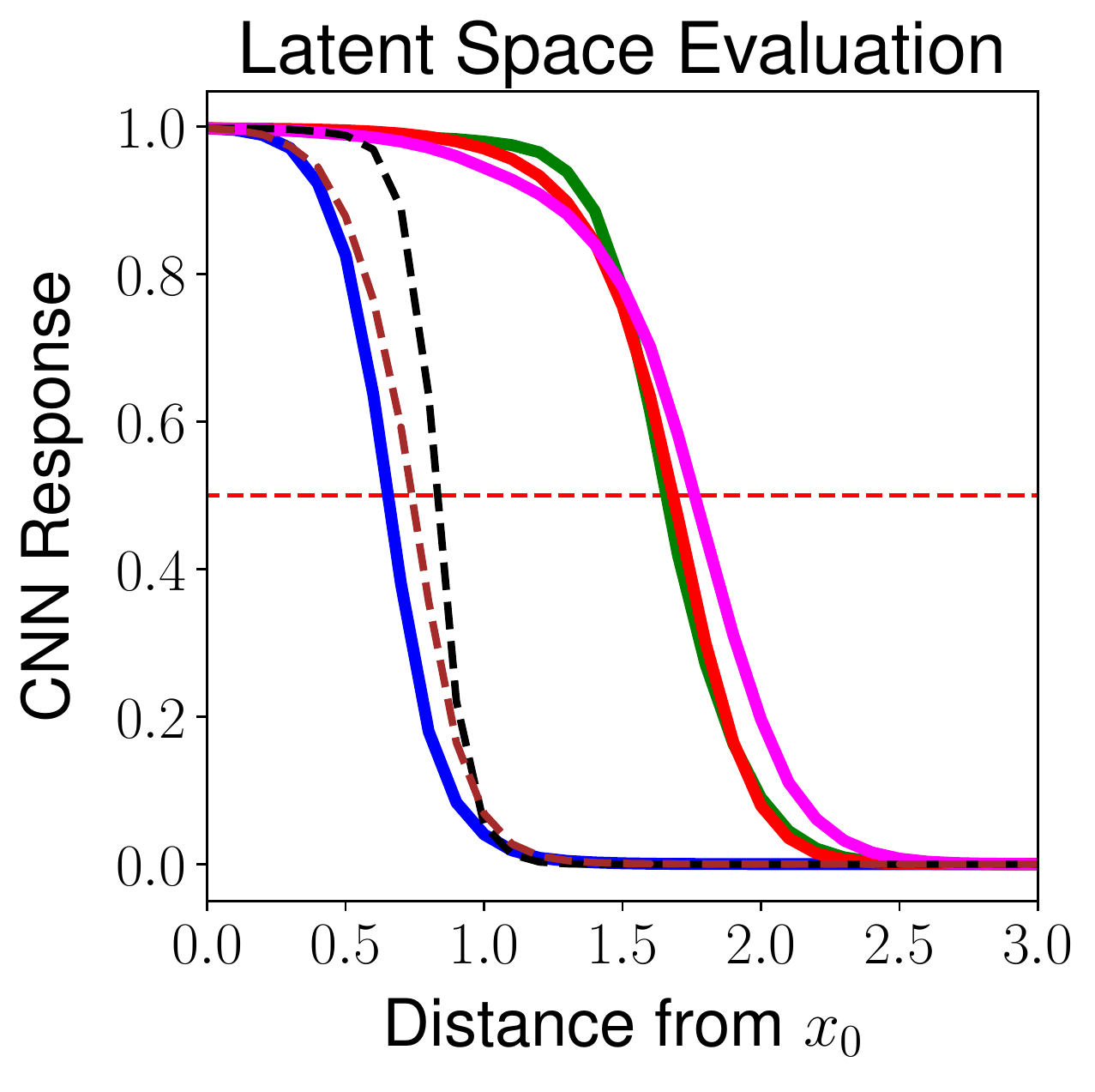}%
        \caption{CNN probabilities moving in direction
        $\latentdirection/\|\latentdirection\|$ from $\x_0$}
        \label{fig:164_latentspace_airis}
    \end{subfigure}%
    \begin{subfigure}[t]{0.45\textwidth}
        \centering
        \includegraphics[width=0.94\textwidth]{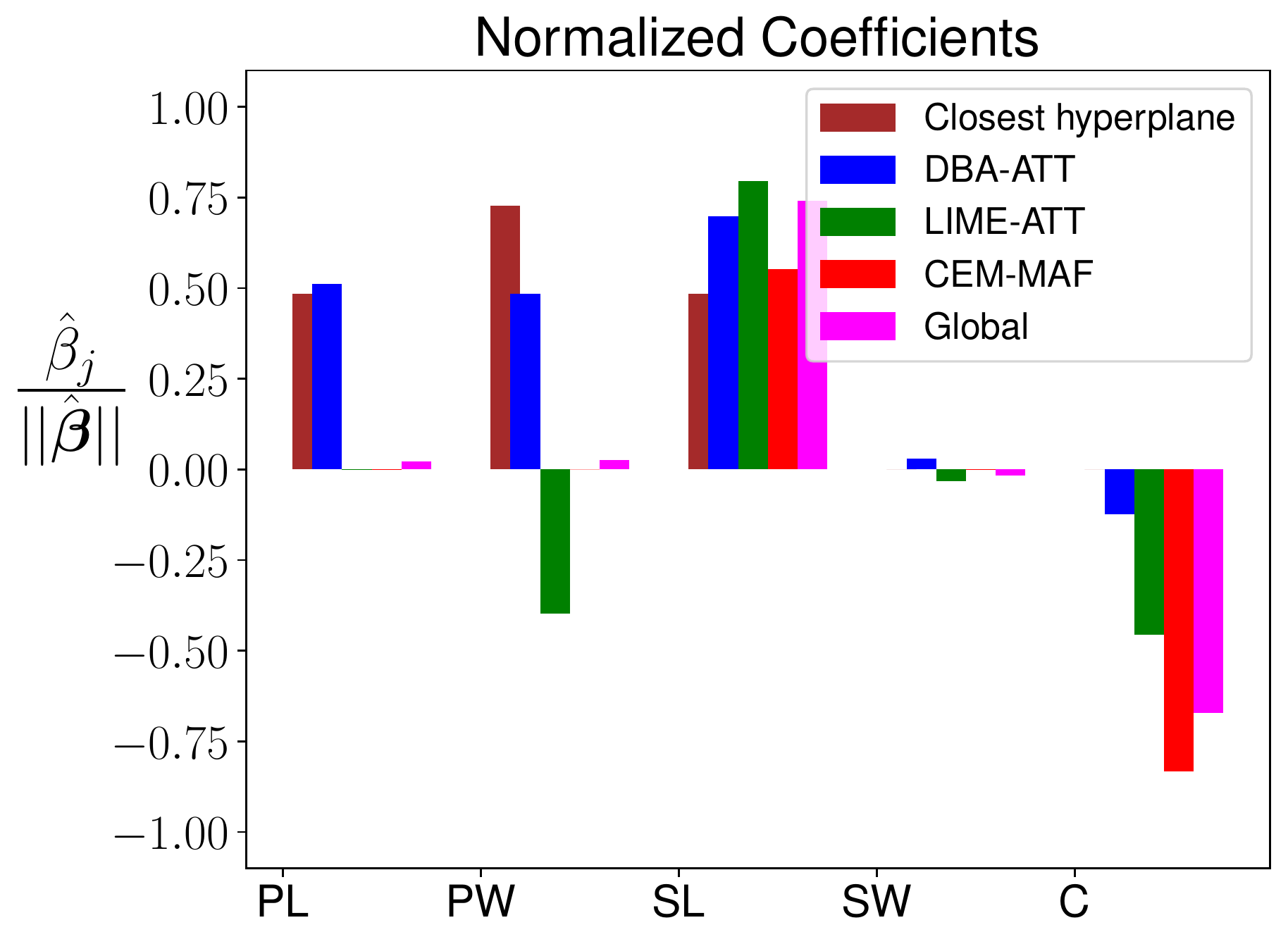}
        \caption{Explanation coefficients $\estbeta$}
        \label{fig:164_coeffs_airis}
    \end{subfigure}
    \caption{Best-case example}
    \label{fig:164_aris}
\end{figure*}

\paragraph{Best-case and Worst-case Examples}

Here we show the results for two more cases, selected from the 50 test
images that were also used in Table~\ref{tab:global_measures_airis}: one
where \dbafull{} performs the worst, in Figure~\ref{fig:141_aris}, and
one where \dbafull{} shows a large advantage over other methods, in
Figure~\ref{fig:164_aris}.

In all 50 cases \dbafull{} found a direction in the latent space that
crossed the decision boundary faster or at least as fast as all other
methods. The worst-case example in Figure~\ref{fig:141_aris} therefore
shows a case where the distance to the decision boundary is
approximately the same for \dbafull{}, \limefull{} and CEM-MAF. Although
the distances to the decision boundary are the same,
Figure~\ref{fig:141_coeffs_airis} shows that the \dbafull{} coefficients
still match slightly better with those of the closest hyperplane.

In the best-case example in Figure~\ref{fig:164_latentspace_airis}, we
see that \dbafull{} succeeds in approximating the closest hyperplane,
while other methods fail to approximate any of the hyperplanes. From
Figure~\ref{fig:164_coeffs_airis} we can conclude that the other methods
average the important features for both hyperplanes in a misleading
manner. For instance, all other methods fail to identify the relevance
of \PL{}, which is an important feature in both hyperplanes but with
opposite sign.

\begin{figure}[htb]
    \centering
    \includegraphics[width = 0.5\textwidth ]{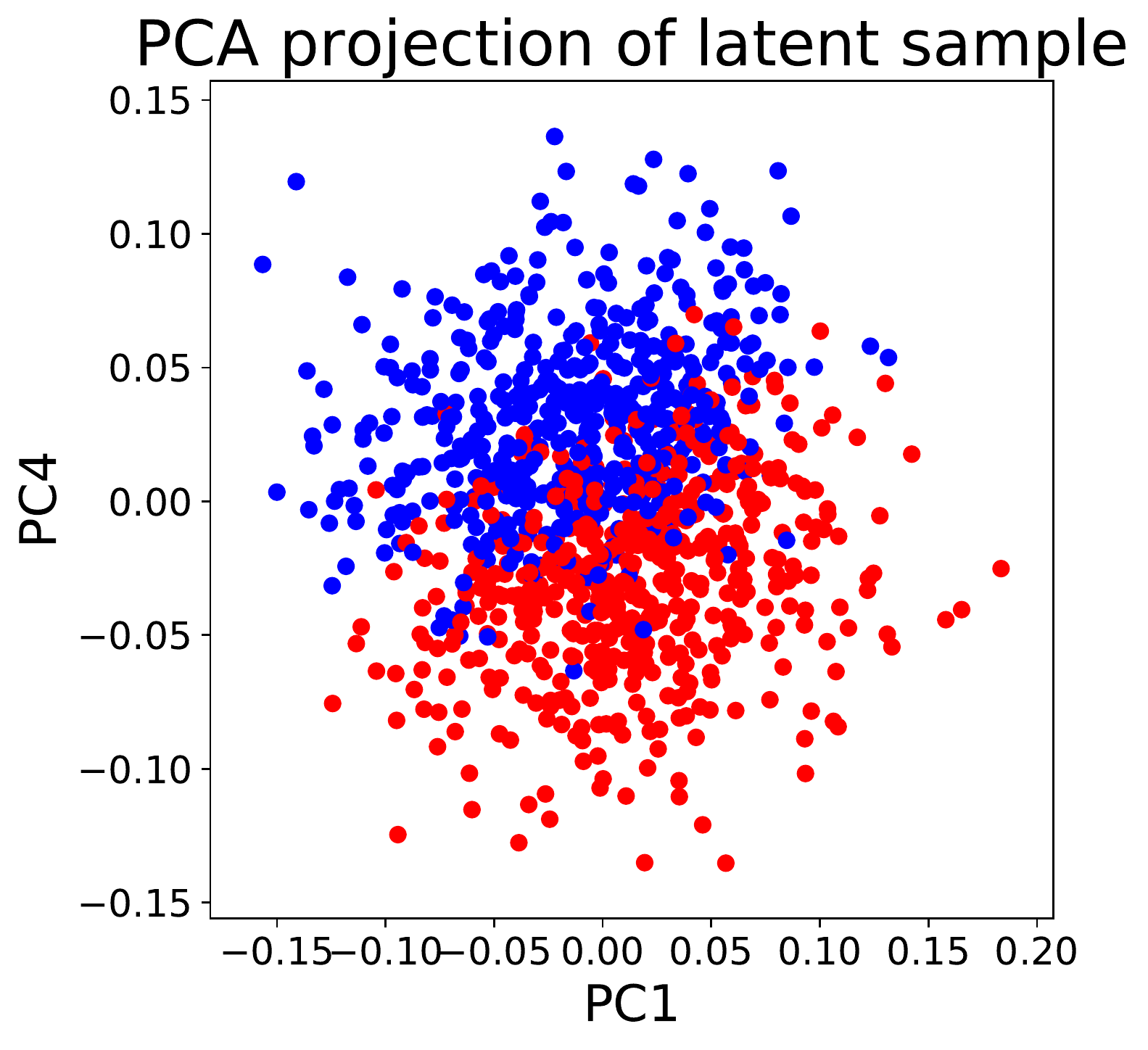}
    \caption{PCA projection of local sample corresponding to the explained example in Section~\ref{sec:AIris} }
    \label{fig:pca_airis}
\end{figure}

\paragraph{PCA}

Figure~\ref{fig:pca_airis} shows the PCA projection of the \dbafull{}
sample $S$ described in Section~\ref{sec:AIris} and corresponding to
Figures~\ref{fig:samples_aires} and \ref{fig:comparison_aIris}. We plot
the components for which the sample is most separable (first and fourth
component) and color by predicted class label of the CNN. This
visualization shows that the simulation step of \dbafull{} creates a
reasonably separable sample on the decision boundary of the CNN, which
can be approximated with high fidelity by a linear surrogate model.

\begin{figure*}[h]
    \centering
    \begin{subfigure}[t]{0.49\textwidth}
        \centering
        \includegraphics[width=1.2\textwidth]{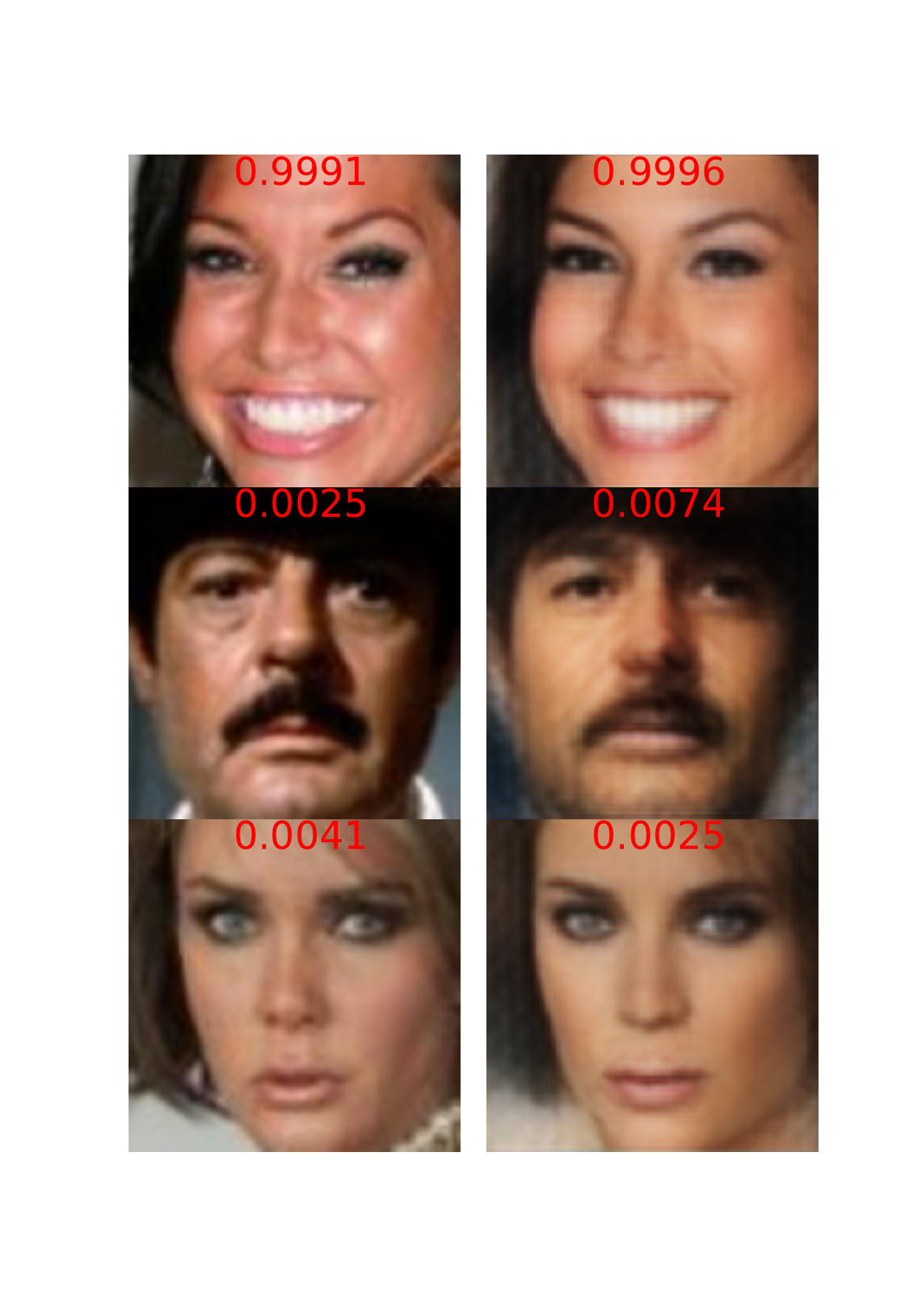}
       \caption{Original vs reconstruction}
        \label{fig:celebareconstructions}
    \end{subfigure}%
    ~ 
    \begin{subfigure}[t]{0.51\textwidth}
        \centering
        \includegraphics[width=2\textwidth]{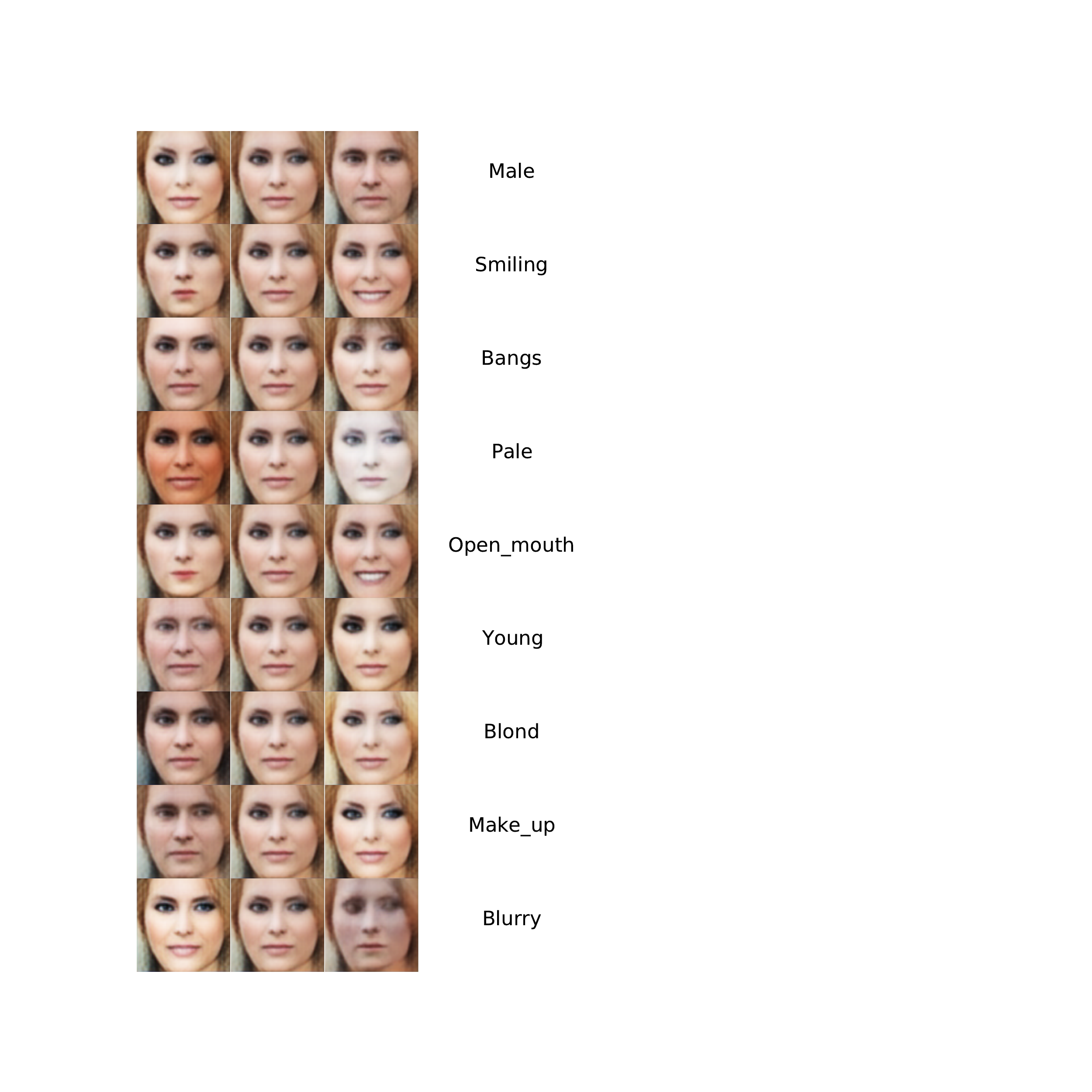}
        \caption{Quality of attribute vectors}
        \label{fig:celebaatt_vecs}
    \end{subfigure}
    \caption{Qualitative evaluation of global mappings. On the left the
    original images and their CNN probabilities are compared with the
    reconstructions produced by the DFC-VAE. On the right the original
    image (middle column) is varied along the direction of each learned
    attribute.}
    \label{fig:celeb_app}
\end{figure*}

\section{CelebA}
\label{app:CelebAdetails}

In this section we provide further details regarding the CelebA
experiment from Section~\ref{sec:celeba}.

\subsection{Global Mappings: CNN, VAE and Annotators}

We first provide details on how we trained the VAE, CNN and annotators
for CelebA. These are the global mappings between spaces in
Figure~\ref{fig:attdba_scheme}.

\subsubsection{VAE}

We use the exact same architecture described in Section~\ref{app:airis}
of this appendix, but we modify the objective function as suggested by
\citet{DFCVAE}. They propose to add an extra term called the deep
feature consistency (DFC) loss, giving a DFC-VAE.
This term involves features of a selected set
of layers from a pre-trained large scale architecture and minimizes the
reconstruction error on these features. We refer to \citet{DFCVAE} for
further details. For our experiment we employ relu1\_1, relu2\_1,
relu3\_1 layers of the pre-trained VGGNet \cite{VGG}. We further use a
latent dimension of 100, batch size 64 and the same optimizer
as in
AIris. We train for 10 epochs with learning rate $10^{-3}$ and monitor
the probability stability from \eqref{eqn:probabilitystability}. 
Figure~\ref{fig:celebareconstructions} illustrates that the
CVAE achieves good reconstruction quality when mapping $\x \mapsto \z
\mapsto \x'$ and that the corresponding CNN probabilities $c(\x)$
and $c(\x')$ are close.

\subsubsection{CNN}

The employed architecture for the CNN consists of 4 convolutional layers
with filter sizes $\{16,32,64,128\}$, kernel size $3 \times 3$ and ReLU
activation. Each of the layers is followed by a 2-dimensional Max
Pooling operation with pool size $2 \times 2$. After the last
convolutional layer a Global Average Pooling operation follows. Next,
there is a dense layer with 64 neurons and ReLU activation which is
followed by Batch Normalization. Finally, the output layer yields the
probability for the smiling female category through a sigmoid activation
function. The network is trained for 10 epochs with learning rate
$10^-3$ and batch size 64. We use the same Adam optimizer as for the
VAE. The resulting CNN achieves train accuracy 93.2\% and test accuracy
91.5\%. 

\subsubsection{Annotators}

\dbafull{} annotators are $L_2$-penalized logistic regression models
which learn the mappings $\latentspace \to \attributespace$. For CelebA,
we train them on a random subset of the training data with 9000 examples
with regularization parameter $\alpha = 0.1$. This shows that we do not
need the entire training set to train the annotators. In
Figure~\ref{fig:celebaatt_vecs} the quality of these vectors can be
observed.

Because of the danger of entanglement discussed in
Section~\ref{sec:discussion}, it is important to verify that the
annotators indeed express the concepts that the user had in mind. This
can be seen in Figure~\ref{fig:celebaatt_vecs}: the middle column always
shows the same example image and the other two columns show the effect
of changing one of the attributes by a unit step in the latent space.
Concretely, we map the input image $\x$ to a latent representation $\z$.
Then add (or subtract) $\esttheta_j/\|\esttheta_j\|$ 
to $\z$ and map back to a corresponding input $\x'$, where
$\esttheta_j$ are the coefficients of the annotator for attribute $j$.
The left column corresponds to decreasing an attribute; the right column
to increasing it. In this case we can see that the attributes
approximately represent the intended concepts. 

\subsection{Details for \dbafull{}, \limefull{} and CEM-MAF}

\paragraph{\dbafull{}}

\begin{sloppypar}
\dbafull{} runs with $k = m = 500$ and the same $\rgrid =
\{0.1,0.2,\ldots,0.9,1,1.5,2,\ldots,9.5,10\}$ as in the AIris
experiment. To speed up the detection step of the algorithm, we do not
select the $k$ closest points from the entire training data, but only
from the subset of 9000 instances that were also used to train the
annotators.
\end{sloppypar}

\paragraph{\limefull{}}

For a general description of \limefull{} see
Appendix~\ref{app:limefull}. For \limefull{} we use $m = 500$ and
$\sigma = 0.75\sqrt{100}$, because the dimension of the latent space is
$100$.

\paragraph{CEM-MAF}

We experiment with CEM-MAF hyperparameters as described in
Appendix~\ref{app:cem_maf}. Reported results are with $\{C , \kappa,
\eta, \nu, \gamma , \mu \} = \{500,5,1,1,100,100\}$ which is quite
similar to the hyperparameters used by \citet{CEM-MAF} in their
experiment in which they apply CEM-MAF to CelebA. The objective
\eqref{eqn:CEM_MAF_loss} is optimized in 1000 epochs of SGD using
polynomial decay for the learning rate with starting value $10^{-2}$ and
power 0.5. With this set-up, CEM-MAF converged to a PN for all of the 30
explained cases.

\begin{figure}[htb]
  \centering
    \begin{subfigure}[t]{\textwidth}
      \centering
      \includegraphics[scale=0.3]{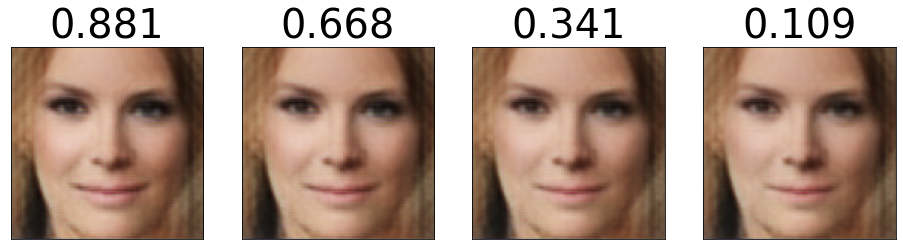}  
      \caption{Transition from class A to class B}
    \end{subfigure}\\
    \begin{subfigure}[t]{\textwidth}
      \centering
      \includegraphics[scale=0.3]{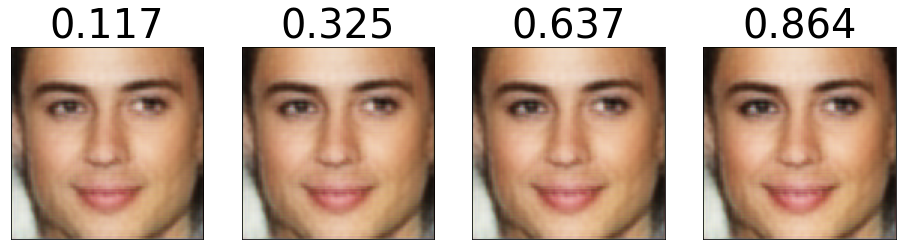}
      \caption{Transition from class B to class A}
    \end{subfigure}
  \caption{The path of $\x_0$ towards the decision boundary of the CNN
  when moving in the latent direction $\latentdirection$ proposed by
  \dbafull{}. In both cases, $\x_0$ is shown in the left-most image. The
  numbers above the images are the predicted probabilities of class~A
  for the CNN.}
  \label{fig:eval_viz}
\end{figure}

\subsection{Additional Result: Visualization of the Path Towards the
Decision Boundary}

We close with an alternative visualization of latent space
evaluations like the one shown in Figure~\ref{fig:24path}.
Figure~\ref{fig:eval_viz} shows two images $\x_0$ and how they change
when we move in the direction $\latentdirection$ produced by \dbafull{}.
We observe in both cases that \dbafull{} has found a very minimal
variation to make the CNN change its the classification.

\end{document}